# Planning by Rewriting

**José Luis Ambite**                                            AMBITE@ISI.EDU
**Craig A. Knoblock**                                        KNOBLOCK@ISI.EDU
*Information Sciences Institute and Department of Computer Science,*
*University of Southern California,*
*4676 Admiralty Way, Marina del Rey, CA 90292, USA*

## Abstract

Domain-independent planning is a hard combinatorial problem. Taking into account plan quality makes the task even more difficult. This article introduces Planning by Rewriting (PbR), a new paradigm for efficient high-quality domain-independent planning. PbR exploits declarative plan-rewriting rules and efficient local search techniques to transform an easy-to-generate, but possibly suboptimal, initial plan into a high-quality plan. In addition to addressing the issues of planning efficiency and plan quality, this framework offers a new anytime planning algorithm. We have implemented this planner and applied it to several existing domains. The experimental results show that the PbR approach provides significant savings in planning effort while generating high-quality plans.

## 1. Introduction

Planning is the process of generating a network of actions, a plan, that achieves a desired goal from an initial state of the world. Many problems of practical importance can be cast as planning problems. Instead of crafting an individual planner to solve each specific problem, a long line of research has focused on constructing domain-independent planning algorithms. Domain-independent planning accepts as input, not only descriptions of the initial state and the goal for each particular problem instance, but also a declarative domain specification, that is, the set of actions that change the properties of the state. Domain-independent planning makes the development of planning algorithms more efficient, allows for software and domain reuse, and facilitates the principled extension of the capabilities of the planner. Unfortunately, domain-independent planning (like most planning problems) is computationally hard (Bylander, 1994; Erol, Nau, & Subrahmanian, 1995; Bäckström & Nebel, 1995). Given the complexity limitations, most of the previous work on domain-independent planning has focused on finding *any* solution plan without careful consideration of plan quality. Usually very simple cost functions, such as the length of the plan, have been used. However, for many practical problems plan quality is crucial. In this paper we present a new planning paradigm, Planning by Rewriting (PbR), that addresses both planning efficiency and plan quality while maintaining the benefits of domain independence. The framework is fully implemented and we present empirical results in several planning domains.

## 1.1 Solution Approach

Two observations guided the present work. The first one is that there are two sources of complexity in planning:

- Satisfiability: the difficulty of finding any solution to the planning problem (regardless of the quality of the solution).

- Optimization: the difficulty of finding the optimal solution under a given cost metric.

For a given domain, each of these facets may contribute differently to the complexity of planning. In particular, there are many domains in which the satisfiability problem is relatively easy and their complexity is dominated by the optimization problem. For example, there may be many plans that would solve the problem, so that finding one is efficient in practice, but the cost of each solution varies greatly, thus finding the optimal one is computationally hard. We will refer to these domains as optimization domains. Some optimization domains of great practical interest are query optimization and manufacturing process planning.[1]

The second observation is that planning problems have a great deal of structure. Plans are a type of graph with strong semantics, determined by both the general properties of planning and each particular domain specification. This structure should and can be exploited to improve the efficiency of the planning process.

Prompted by the previous observations, we developed a novel approach for efficient planning in optimization domains: Planning by Rewriting (PbR). The framework works in two phases:

1. Generate an initial solution plan. Recall that in optimization domains this is efficient. However, the quality of this initial plan may be far from optimal.

2. Iteratively rewrite the current solution plan improving its quality using a set of declarative plan-rewriting rules, until either an acceptable solution is found or a resource limit is reached.

As motivation, consider the optimization domains of distributed query processing and manufacturing process planning.[2] Distributed query processing (Yu & Chang, 1984) involves generating a plan that efficiently computes a user query from data that resides at different nodes in a network. This query plan is composed of data retrieval actions at diverse information sources and operations on this data (such as those of the relational algebra: join, selection, etc). Some systems use a general-purpose planner to solve this problem (Knoblock, 1996). In this domain it is easy to construct an initial plan (any parse of the query suffices) and then transform it using a gradient-descent search to reduce its cost. The plan transformations exploit the commutative and associative properties of the (relational algebra) operators, and facts such as that when a group of operators can be executed together at a remote information source it is generally more efficient to do so. Figure 1 shows some sample transformations. `Simple-join-swap` transforms two join trees according to the commutative and associative properties of the join operator. `Remote-join-eval` executes a join of two subqueries at a remote source, if the source is able to do so.

---

1. Interestingly, one of the most widely studied planning domains, the Blocks World, also has this property.
2. These domains are analyzed in Section 4. Graphical examples of the rewriting process appear in Figure 30 for query planning and in Figure 21 for manufacturing process planning. The reader may want to consult those figures even if not all details can be explained at this point.

---

**Simple-Join-Swap:**

$retrieve(Q1, Source1) \bowtie [retrieve(Q2, Source2) \bowtie retrieve(Q3, Source3)] \Leftrightarrow$
$retrieve(Q2, Source2) \bowtie [retrieve(Q1, Source1) \bowtie retrieve(Q3, Source3)]$

**Remote-Join-Eval:**

$(retrieve(Q1, Source) \bowtie retrieve(Q2, Source)) \land capability(Source, join)$
$\Rightarrow retrieve(Q1 \bowtie Q2, Source)$

---

Figure 1: Transformations in Query Planning

In manufacturing, the problem is to find an economical plan of machining operations that implement the desired features of a design. In a feature-based approach (Nau, Gupta, & Regli, 1995), it is possible to enumerate the actions involved in building a piece by analyzing its CAD model. It is more difficult to find an ordering of the operations and the setups that optimize the machining cost. However, similar to query planning, it is possible to incrementally transform a (possibly inefficient) initial plan. Often, the order of actions does not affect the design goal, only the quality of the plan, thus many actions can commute. Also, it is important to minimize the number of setups because fixing a piece on a machine is a rather time consuming operation. Interestingly, such grouping of machining operations on a setup is analogous to evaluating a subquery at a remote information source.

As suggested by these examples, there are many problems that combine the characteristics of traditional planning satisfiability with quality optimization. For these domains there often exist natural transformations that may be used to efficiently obtain high-quality plans by iterative rewriting. Planning by Rewriting provides a domain-independent framework that allows plan transformations to be conveniently specified as declarative plan-rewriting rules and facilitates the exploration of efficient (local) search techniques.

## 1.2 Advantages of Planning by Rewriting

There are several advantages to the planning style that PbR introduces. First, PbR is a declarative domain-independent framework. This facilitates the specification of planning domains, their evolution, and the principled extension of the planner with new capabilities. Moreover, the declarative rewriting rule language provides a natural and convenient mechanism to specify complex plan transformations.

Second, PbR accepts sophisticated quality measures because it operates on complete plans. Most previous planning approaches either have not addressed quality issues or have very simple quality measures, such as the number of steps in the plan, because only partial plans are available during the planning process. In general, a partial plan cannot offer enough information to evaluate a complex cost metric and/or guide the planning search effectively.

Third, PbR can use local search methods that have been remarkably successful in scaling to large problems (Aarts & Lenstra, 1997).[3] By using local search techniques, high-quality plans can be efficiently generated. Fourth, the search occurs in the space of solution plans, which is generally much smaller than the space of partial plans explored by planners based on refinement search.

Fifth, our framework yields an anytime planning algorithm (Dean & Boddy, 1988). The planner always has a solution to offer at any point in its computation (modulo the initial plan generation that needs to be fast). This is a clear advantage over traditional planning approaches, which must run to completion before producing a solution. Thus, our system allows the possibility of trading off planning effort and plan quality. For example, in query planning the quality of a plan is its execution time and it may not make sense to keep planning if the cost of the current plan is small enough, even if a cheaper one could be found. Further discussion and concrete examples of these advantages are given throughout the following sections.

## 1.3 Contributions

The main contribution of this paper is the development of Planning by Rewriting, a novel domain-independent paradigm for efficient high-quality planning. First, we define a language of declarative plan-rewriting rules and present the algorithms for domain-independent plan rewriting. The rewriting rules provide a natural and convenient mechanism to specify complex plan transformations. Our techniques for plan rewriting generalize traditional graph rewriting. Graph rewriting rules need to specify in the rule consequent the complete embedding of the replacement subplan. We introduce the novel class of partially-specified plan-rewriting rules that relax that restriction. By taking advantage of the semantics of planning, this embedding can be automatically computed. A single partially-specified rule can concisely represent a great number of fully-specified rules. These rules are also easier to write and understand than their fully-specified counterparts. Second, we adapt local search techniques, such as gradient descent, to efficiently explore the space of plan rewritings and optimize plan quality. Finally, we demonstrate empirically the usefulness of the PbR approach in several planning domains.

## 1.4 Outline

The remainder of this paper is structured as follows. Section 2 provides background on planning, rewriting, and local search, some of the fields upon which PbR builds. Section 3 presents the basic framework of Planning by Rewriting as a domain-independent approach to local search. This section describes in detail plan rewriting and our declarative rewriting rule language. Section 4 describes several application domains and shows experimental results comparing PbR with other planners. Section 5 reviews related work. Finally, Section 6 summarizes the contributions of the paper and discusses future work.

---

3. Although the space of rewritings can be explored by complete search methods, in the application domains we have analyzed the search space is very large and our experience suggests that local search is more appropriate. However, to what extent complete search methods are useful in a Planning by Rewriting framework remains an open issue. In this paper we focus on local search.

## 2. Preliminaries: Planning, Rewriting, and Local Search

The framework of Planning by Rewriting arises as the confluence of several areas of research, namely, artificial intelligence planning algorithms, graph rewriting, and local search techniques. In this section we give some background on these areas and explain how they relate to PbR.

### 2.1 AI Planning

We assume that the reader is familiar with classical AI planning, but in this section we will highlight the main concepts and relate them to the PbR framework. Weld (1994, 1999) and Russell & Norvig (1995) provide excellent introductions to AI planning.

PbR follows the classical AI planning representation of actions that transform a state. The state is a set of ground propositions understood as a conjunctive formula. PbR, as most AI planners, follows the *Closed World Assumption*, that is, if a proposition is not explicitly mentioned in the state it is assumed to be false, similarly to the *negation as failure* semantics of logic programming. The propositions of the state are modified, asserted or negated, by the actions in the domain. The actions of a domain are specified by operator schemas.

An operator schema consists of two logical formulas: the precondition, which defines the conditions under which the operator may be applied, and the postcondition, which specifies the changes on the state effected by the operator. Propositions not mentioned in the postcondition are assumed not to change during the application of the operator. This type of representation was initially introduced in the STRIPS system (Fikes & Nilsson, 1971). The language for the operators in PbR is the same as in Sage (Knoblock, 1995, 1994b), which is an extension of UCPOP (Penberthy & Weld, 1992). The operator description language in PbR accepts arbitrary function-free first-order formulas in the preconditions of the operators, and conditional and universally quantified effects (but no disjunctive effects). In addition, the operators can specify the resources they use. Sage and PbR address unit non-consumable resources. These resources are fully acquired by an operator until the completion of its action and then released to be reused.

Figure 2 shows a sample operator schema specification for a simple Blocks World domain,[4] in the representation accepted by PbR. This domain has two actions: `stack`, which puts one block on top of another, and `unstack`, which places a block on the table.[5] The state is described by two predicates: `(on ?x ?y)`[6] denotes that a block `?x` is on top of another block `?y` (or on the Table), and `(clear ?x)` denotes that a `?x` block does not have any other block on top of it.

An example of a more complex operator from a process manufacturing domain is shown in Figure 3. This operator describes the behavior of a punch, which is a machine used to make holes in parts. The punch operation requires that there is an available clamp at the machine and that the orientation and width of the hole is appropriate for using the punch. After executing the operation the part will have the desired hole but it will also have a

---

4. To illustrate the basic concepts in planning, we will use examples from a simple Blocks World domain. The reader will find a "real-world" application of planning techniques, query planning, in Section 4.4.
5. `(stack ?x ?y ?z)` can be read as stack the block `?x` on top of block `?y` from `?z`.
   `(unstack ?x ?y)` can be read as lift block `?x` from the top of block `?y` and put it on the Table.
6. By convention, variables are preceded by a question mark symbol (?), as in `?x`.

```
(define (operator STACK)                    (define (operator UNSTACK)
 :parameters (?X ?Y ?Z)                      :parameters (?X ?Y)
 :precondition                               :precondition
  (:and (on ?X ?Z) (clear ?X) (clear ?Y)     (:and (on ?X ?Y) (clear ?X) (:neq ?X ?Y)
     (:neq ?Y ?Z) (:neq ?X ?Z) (:neq ?X ?Y)      (:neq ?X Table) (:neq ?Y Table))
     (:neq ?X Table) (:neq ?Y Table))       :effect (:and (on ?X Table) (clear ?Y)
 :effect (:and (on ?X ?Y) (:not (on ?X ?Z))         (:not (on ?X ?Y))))
         (clear ?Z) (:not (clear ?Y))))
```

Figure 2: Blocks World Operators

```
(define (operator PUNCH)
 :parameters (?x ?width ?orientation)
 :resources ((machine PUNCH) (is-object ?x))
 :precondition (:and (is-object ?x)
                     (is-punchable ?x ?width ?orientation)
                     (has-clamp PUNCH))
 :effect (:and (:forall (?surf) (:when (:neq ?surf ROUGH)
                                       (:not (surface-condition ?x ?surf))))
               (surface-condition ?x ROUGH)
               (has-hole ?x ?width ?orientation)))
```

Figure 3: Manufacturing Operator

rough surface.[7] Note the specification on the resources slot. Declaring (machine PUNCH) as a resource enforces that no other operator can use the punch concurrently. Similarly, declaring the part, (is-object ?x), as a resource means that only one operation at a time can be performed on the object. Further examples of operator specifications appear in Figures 18, 19, and 28.

A plan in PbR is represented by a graph, in the spirit of partial-order causal-link planners (POCL) such as UCPOP (Penberthy & Weld, 1992). The nodes are plan steps, that is, instantiated domain operators. The edges specify a temporal ordering relation among steps imposed by causal links and ordering constraints. A causal link is a record of how a proposition is established in a plan. This record contains the proposition (sometimes also called a condition), a producer step, and a consumer step. The producer is a step in the plan that asserts the proposition, that is, the proposition is one of its effects. The consumer is a step that needs that proposition, that is, the proposition is one of its preconditions. By causality, the producer must precede the consumer.

The ordering constraints are needed to ensure that the plan is consistent. They arise from resolving operator threats and resource conflicts. An operator threat occurs when a step that negates the condition of a causal link can be ordered between the producer and the consumer steps of the causal link. To prevent this situation, which makes the plan inconsistent, POCL planners order the threatening step either before the producer (demotion) or after the consumer (promotion) by posting the appropriate ordering constraints. For the unit non-consumable resources we considered, steps requiring the same resource have to be sequentially ordered, and such a chain of ordering constraints will appear in the plan.

---

7. This operator uses an idiom combining universal quantification and negated conditional effects to enforce that the attribute surface-condition of a part is single-valued.

An example of a plan in the Blocks World using this graph representation is given in Figure 4. This plan transforms an initial state consisting of two towers: C on A, A on the Table, B on D, and D on the Table; to the final state consisting of one tower: A on B, B on C, C on D, and D on the Table. The initial state is represented as step 0 with no preconditions and all the propositions of the initial state as postconditions. Similarly, the goal state is represented as a step goal with no postconditions and the goal formula as the precondition. The plan achieves the goal by using two unstack steps to disassemble the two initial towers and then using three stack steps to build the desired tower. The causal links are shown as solid arrows and the ordering constraints as dashed arrows. The additional effects of a step that are not used in causal links, sometimes called side effects, are shown after each step pointed by thin dashed arrows. Negated propositions are preceded by ¬. Note the need for the ordering link between the steps 2, stack(B C Table), and 3, stack(A B Table). If step 3 could be ordered concurrently or before step 2, it would negate the precondition clear(B) of step 2, making the plan inconsistent. A similar situation occurs between steps 1 and 2 where another ordering link is introduced.

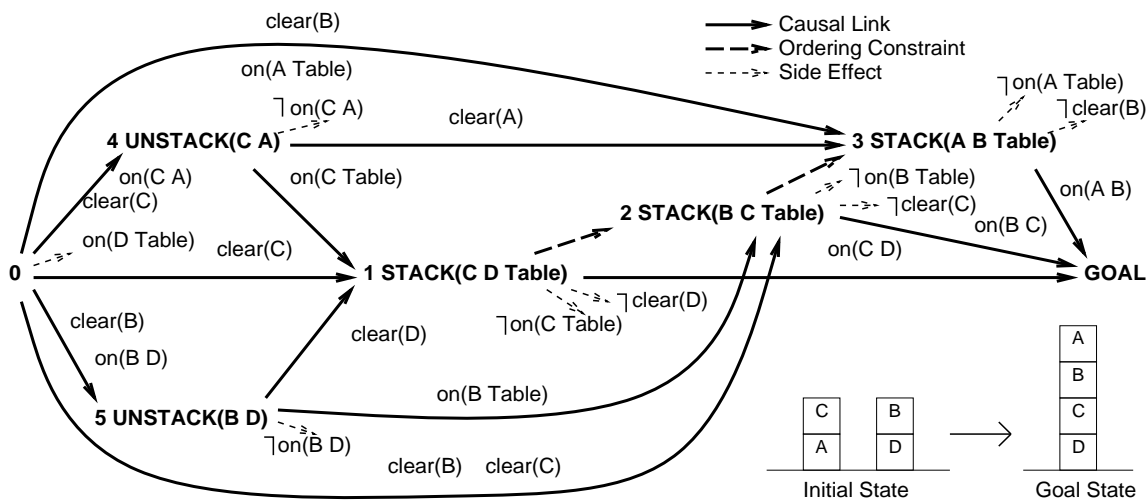

Figure 4: Sample Plan in the Blocks World Domain

## 2.2 Rewriting

Plan rewriting in PbR is related to term and graph rewriting. Term rewriting originated in the context of equational theories and reduction to normal forms as an effective way to perform deduction (Avenhaus & Madlener, 1990; Baader & Nipkow, 1998). A rewrite system is specified as a set of rules. Each rule corresponds to a preferred direction of an equivalence theorem. The main issue in term rewriting systems is convergence, that is, if two arbitrary terms can be rewritten in a finite number of steps into a unique normal form. In PbR two plans are considered "equivalent" if they are solutions to the same problem, although they may differ on their cost or operators (that is, they are "equivalent" with respect to "satisfiability" as introduced above). However, we are not interested in using the rewriting rules to prove such "equivalence". Instead, our framework uses the rewriting rules to explore the space of solution plans.

Graph rewriting, akin to term rewriting, refers to the process of replacing a subgraph of a given graph, when some conditions are satisfied, by another subgraph. Graph rewriting has found broad applications, such as very high-level programming languages, database data description and query languages, etc. Schürr (1997) presents a good survey. The main drawback of general graph rewriting is its complexity. Because graph matching can be reduced to (sub)graph isomorphism the problem is NP-complete. Nevertheless, under some restrictions graph rewriting can be performed efficiently (Dorr, 1995).

Planning by Rewriting adapts general graph rewriting to the semantics of partial-order planning with a STRIPS-like operator representation. A plan-rewriting rule in PbR specifies the replacement, under certain conditions, of a subplan by another subplan. However, in our formalism the rule does not need to specify the completely detailed embedding of the consequent as in graph rewriting systems. The consistent embeddings of the rule consequent, with the generation of edges if necessary, are automatically computed according to the semantics of partial-order planning. Our algorithm ensures that the rewritten plans always remain valid (Section 3.1.3). The plan-rewriting rules are intended to explore the space of solution plans to reach high-quality plans.

## 2.3 Local Search in Combinatorial Optimization

PbR is inspired by the local search techniques used in combinatorial optimization. An instance of a combinatorial optimization problem consists of a set of feasible solutions and a cost function over the solutions. The problem consists of finding a solution with the optimal cost among all feasible solutions. Generally the problems addressed are computationally intractable, thus approximation algorithms have to be used. One class of approximation algorithms that have been surprisingly successful in spite of their simplicity are local search methods (Aarts & Lenstra, 1997; Papadimitriou & Steiglitz, 1982).

Local search is based on the concept of a neighborhood. A neighborhood of a solution $p$ is a set of solutions that are in some sense close to $p$, for example because they can be easily computed from $p$ or because they share a significant amount of structure with $p$. The neighborhood generating function may, or may not, be able to generate the global optima solution. When the neighborhood function can generate the global optima, starting from any initial feasible point, it is called *exact* (Papadimitriou & Steiglitz, 1982, page 10).

Local search can be seen as a walk on a directed graph whose vertices are solutions points and whose arcs connect neighboring points. The neighborhood generating function determines the properties of this graph. In particular, if the graph is disconnected, then the neighborhood is not exact since there exist feasible points that would lead to local optima but not the global optima. In PbR the points are solution plans and the neighbors of a plan are the plans generated by the application of a set of declarative plan rewriting rules.

The basic version of local search is *iterative improvement*. Iterative improvement starts with an initial solution and searches a neighborhood of the solution for a lower cost solution. If such a solution is found, it replaces the current solution and the search continues.

Otherwise, the algorithm returns a locally optimal solution. Figure 5(a) shows a graphical depiction of basic iterative improvement. There are several variations of this basic algorithm. *First improvement* generates the neighborhood incrementally and selects the first solution of better cost than the current one. *Best improvement* generates the complete neighborhood and selects the best solution within this neighborhood.

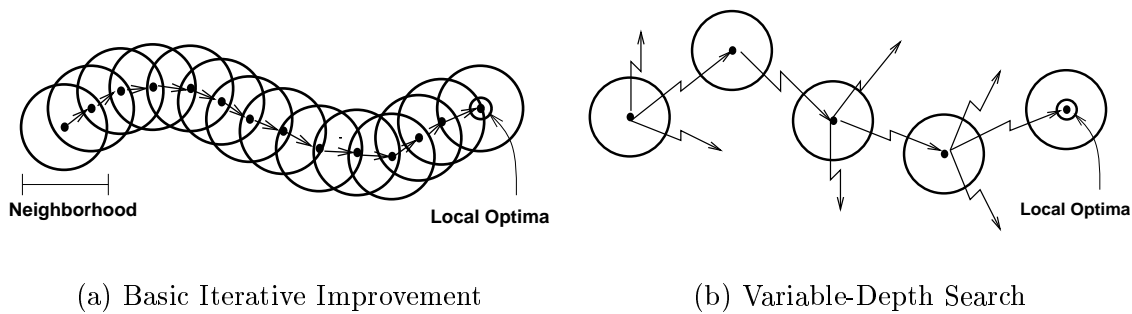

(a) Basic Iterative Improvement        (b) Variable-Depth Search

Figure 5: Local Search

Basic iterative improvement obtains local optima, not necessarily the global optimum. One way to improve the quality of the solution is to restart the search from several initial points and choose the best of the local optima reached from them. More advanced algorithms, such as variable-depth search, simulated annealing and tabu search, attempt to minimize the probability of being stuck in a low-quality local optimum.

*Variable-depth search* is based on applying a sequence of steps as opposed to only one step at each iteration. Moreover, the length of the sequence may change from iteration to iteration. In this way the system overcomes small cost increases if eventually they lead to strong cost reductions. Figure 5(b) shows a graphical depiction of variable-depth search.

*Simulated annealing* (Kirkpatrick, Gelatt, & Vecchi, 1983) selects the next point randomly. If a lower cost solution is chosen, it is selected. If a solution of a higher cost is chosen, it is still selected with some probability. This probability is decreased as the algorithm progresses (analogously to the temperature in physical annealing). The function that governs the behavior of the acceptance probability is called the cooling schedule. It can be proven that simulated annealing converges asymptotically to the optimal solution. Unfortunately, such convergence requires exponential time. So, in practice, simulated annealing is used with faster cooling schedules (not guaranteed to converge to the optimal) and thus it behaves like an approximation algorithm.

*Tabu search* (Glover, 1989) can also accept cost-increasing neighbors. The next solution is a randomly chosen legal neighbor even if its cost is worse than the current solution. A neighbor is legal if it is not in a limited-size tabu list. The dynamically updated tabu list prevents some solution points from being considered for some period of time. The intuition is that if we decide to consider a solution of a higher cost at least it should lie in an unexplored part of the space. This mechanism forces the exploration of the solution space out of local minima.

Finally, we should stress that the appeal of local search relies on its simplicity and good average-case behavior. As could be expected, there are a number of negative worst-case results. For example, in the traveling salesman problem it is known that exact neighborhoods, that do not depend on the problem instance, must have exponential size (Savage, Weiner, & Bagchi, 1976). Moreover, an improving move in these neighborhoods cannot be found in polynomial time unless P = NP (Papadimitriou & Steiglitz, 1977). Nevertheless, the best approximation algorithm for the traveling salesman problem is a local search algorithm (Johnson, 1990).

## 3. Planning by Rewriting as Local Search

Planning by Rewriting can be viewed as a domain-independent framework for local search. PbR accepts arbitrary domain specifications, declarative plan-rewriting rules that generate the neighborhood of a plan, and arbitrary (local) search methods. Therefore, assuming that a given combinatorial problem can be encoded as a planning problem, PbR can take it as input and experiment with different neighborhoods and search methods.

We will describe the main issues in Planning by Rewriting as an instantiation of the local search idea typical of combinatorial optimization algorithms:

- *Selection of an initial feasible point*: In PbR this phase consists of efficiently generating an initial solution plan.

- *Generation of a local neighborhood*: In PbR the neighborhood of a plan is the set of plans obtained from the application of a set of declarative plan-rewriting rules.

- *Cost function to minimize*: This is the measure of plan quality that the planner is optimizing. The plan quality function can range from a simple domain-independent cost metric, such as the number of steps, to more complex domain-specific ones, such as the query evaluation cost or the total manufacturing time for a set of parts.

- *Selection of the next point*: In PbR, this consists of deciding which solution plan to consider next. This choice determines how the global space will be explored and has a significant impact on the efficiency of planning. A variety of local search strategies can be used in PbR, such as steepest descent, simulated annealing, etc. Which search method yields the best results may be domain or problem specific.

In the following subsections we expand on these issues. First, we discuss the use of declarative rewriting rules to generate a local neighborhood of a plan, which constitutes the main contribution of this paper. We present the syntax and semantics of the rules, the plan-rewriting algorithm, the formal properties and a complexity analysis of plan rewriting, and a rule taxonomy. Second, we address the selection of the next plan and the associated search techniques for plan optimization. Third, we discuss the measures of plan quality. Finally, we describe some approaches for initial plan generation.

### 3.1 Local Neighborhood Generation: Plan-Rewriting Rules

The neighborhood of a solution plan is generated by the application of a set of declarative plan-rewriting rules. These rules embody the domain-specific knowledge about what transformations of a solution plan are likely to result in higher-quality solutions. The application of a given rule may produce one or several rewritten plans or fail to produce a plan, but the rewritten plans are guaranteed to be valid solutions. First, we describe the syntax and semantics of the rules. Second, we introduce two approaches to rule specification. Third, we present the rewriting algorithm, its formal properties, and the complexity of plan rewriting. Finally, we present a taxonomy of plan-rewriting rules.

### 3.1.1 PLAN-REWRITING RULES: SYNTAX AND SEMANTICS

First, we introduce the rule syntax and semantics through some examples. Then, we provide a formal description. A plan-rewriting rule has three components: (1) the antecedent (`:if` field) specifies a subplan to be matched; (2) the `:replace` field identifies the subplan that is going to be removed, a subset of steps and links of the antecedent; (3) the `:with` field specifies the replacement subplan. Figure 6 shows two rewriting rules for the Blocks World domain introduced in Figure 2. Intuitively, the rule `avoid-move-twice` says that, whenever possible, it is better to stack a block on top of another directly, rather than first moving it to the table. This situation occurs in plans generated by the simple algorithm that first puts all blocks on the table and then build the desired towers, such as the plan in Figure 4. The rule `avoid-undo` says that the actions of moving a block to the table and back to its original position cancel each other and both could be removed from a plan.

```
(define-rule :name avoid-move-twice              (define-rule :name avoid-undo
 :if (:operators ((?n1 (unstack ?b1 ?b2))         :if (:operators
                  (?n2 (stack ?b1 ?b3 Table)))          ((?n1 (unstack ?b1 ?b2))
       :links (?n1 (on ?b1 Table) ?n2)                  (?n2 (stack ?b1 ?b2 Table)))
       :constraints ((possibly-adjacent ?n1 ?n2)        :constraints
                     (:neq ?b2 ?b3)))                        ((possibly-adjacent ?n1 ?n2))
 :replace (:operators (?n1 ?n2))                  :replace (:operators (?n1 ?n2))
 :with (:operators (?n3 (stack ?b1 ?b3 ?b2))))    :with NIL))
```

Figure 6: Blocks World Rewriting Rules

A rule for the manufacturing domain of (Minton, 1988b) is shown in Figure 7. This domain and additional rewriting rules are described in detail in Section 4.1. The rule states that if a plan includes two consecutive punching operations in order to make holes in two different objects, but another machine, a drill-press, is also available, the plan quality may be improved by replacing one of the punch operations with the drill-press. In this domain the plan quality is the (parallel) time to manufacture all parts. This rule helps to parallelize the plan and thus improve the plan quality.

```
(define-rule :name punch-by-drill-press
  :if (:operators ((?n1 (punch ?o1 ?width1 ?orientation1))
                   (?n2 (punch ?o2 ?width2 ?orientation2)))
        :links (?n1 ?n2)
        :constraints ((:neq ?o1 ?o2)
                      (possibly-adjacent ?n1 ?n2)))
  :replace (:operators (?n1))
  :with (:operators (?n3 (drill-press ?o1 ?width1 ?orientation1))))
```

Figure 7: Manufacturing Process Planning Rewriting Rule

The plan-rewriting rule syntax is described by the BNF specification given in Figure 8. This BNF generates rules that follow the template shown in Figure 9. Next, we describe the semantics of the three components of a rule (`:if`, `:replace`, and `:with` fields) in detail.

```
<rule> ::= (define-rule :name <name>
                         :if (<graph-spec-with-constraints>)
                         :replace (<graph-spec>)
                         :with (<graph-spec>))
<graph-spec-with-constraints> ::= {<graph-spec>}
                                  {:constraints (<constraints>)}
<graph-spec> ::= {:operators (<nodes>)}
                 {:links (<edges>)} | NIL
<nodes> ::= <node> | <node> <nodes>
<edges> ::= <edge> | <edge> <edges>
<constraints> ::= <constraint> | <constraint> <constraints>
<node> ::= (<node-var> {<node-predicate>} {:resource})
<edge> ::= (<node-var> <node-var>) |
           (<node-var> <edge-predicate> <node-var>) |
           (<node-var> :threat <node-var>)
<constraint> ::= <interpreted-predicate> |
                 (:neq <pred-var> <pred-var>)
<node-var> ∩ <pred-var> = ∅, {} = optional, | = alternative
```

Figure 8: BNF for the Rewriting Rules

```
(define-rule :name <rule-name>
  :if (:operators ((<nv> <np> {:resource}) ...)
       :links ((<nv> {<lp>|:threat} <nv>) ...)
       :constraints (<ip> ...))
  :replace (:operators (<nv> ...)
            :links ((<nv> {<lp>|:threat} <nv>) ...))
  :with (:operators ((<nv> <np> {:resource}) ...)
         :links ((<nv> {<lp>} <nv>) ...)))

<nv> = node variable, <np> = node predicate, {} = optional
<lp> = causal link predicate, <ip> = interpreted predicate,  | = alternative
```

Figure 9: Rewriting Rule Template

The antecedent, the `:if` field, specifies a subplan to be matched against the current plan. The graph structure of the subplan is defined in the `:operators` and `:links` fields. The `:operators` field specifies the nodes (operators) of the graph and the `:links` field specifies the edges (causal and ordering links). Finally, the `:constraints` field specifies a set of constraints that the operators and links must satisfy.

The `:operators` field consists of a list of node variable and node predicate pairs. The step number of those steps in the plan that match the given node predicate would be correspondingly bound to the node variable. The node predicate can be interpreted in two ways: as the step action, or as a resource used by the step. For example, the node specification (`?n2 (stack ?b1 ?b3 Table)`) in the antecedent of `avoid-move-twice` in Figure 6 shows a node predicate that denotes a step action. This node specification will collect tuples, composed of step number `?n2` and blocks `?b1` and `?b3`, obtained by matching steps whose action is a `stack` of a block `?b1` that is on the `Table` and it is moved on top of another block `?b3`. This node specification applied to the plan in Figure 4 would result in three matches: (1 C D), (2 B C), and (3 A B), for the variables (`?n2 ?b1 ?b3`) respectively. If the optional keyword `:resource` is present, the node predicate is interpreted as one of the resources used by a plan step, as opposed to describing a step action. An example of a rule that matches against the resources of an operator is given in Figure 10, where the node specification (`?n1 (machine ?x) :resource`) will match all steps that use a resource of type `machine` and collect pairs of step number `?n1` and machine object `?x`.

```
(define-rule :name resource-swap
  :if (:operators ((?n1 (machine ?x) :resource)
                   (?n2 (machine ?x) :resource))
       :links ((?n1 :threat ?n2)))
  :replace (:links (?n1 ?n2))
  :with (:links (?n2 ?n1)))
```

Figure 10: Resource-Swap Rewriting Rule

The `:links` field consists of a list of link specifications. Our language admits link specifications of three types. The first type is specified as a pair of node variables. For example, (`?n1 ?n2`) in Figure 7. This specification matches any temporal ordering link in the plan, regardless if it was imposed by causal links or by the resolution of threats.

The second type of link specification matches causal links. Causal links are specified as triples composed of a producer step node variable, an edge predicate, and a consumer step node variable. The semantics of a causal link is that the producer step asserts in its effects the predicate, which in turn is needed in the preconditions of the consumer step. For example, the link specification (`?n1 (on ?b1 Table) ?n2`) in Figure 6 matches steps `?n1` that put a block `?b1` on the `Table` and steps `?n2` that subsequently pick up this block. That link specification applied to the plan in Figure 4 would result in the matches: (4 C 1) and (5 B 2), for the variables (`?n1 ?b1 ?n2`).

The third type of link specification matches ordering links originating from the resolution of threats (coming either from resource conflicts or from operator conflicts). These links are selected by using the keyword `:threat` in the place of a condition. For example, the `resource-swap` rule in Figure 10 uses the link specification (`?n1 :threat ?n2`) to ensure that only steps that are ordered because they are involved in a threat situation are matched. This helps to identify which are the "critical" steps that do not have any other reasons (i.e. causal links) to be in such order, and therefore this rule may attempt to reorder them. This is useful when the plan quality depends on the degree of parallelism in the plan as a different ordering may help to parallelize the plan. Recall that threats can be solved either by promotion or demotion, so the reverse ordering may also produce a valid plan, which is often the case when the conflict is among resources as in the rule in Figure 10.

Interpreted predicates, built-in and user-defined, can be specified in the `:constraints` field. These predicates are implemented programmatically as opposed to being obtained by matching against components from the plan. The built-in predicates currently implemented are inequality[8] (`:neq`), comparison (`< <= > >=`), and arithmetic (`+ - * /`) predicates. The user can also add arbitrary predicates and their corresponding programmatic implementa-tions. The interpreted predicates may act as filters on the previous variables or introduce new variables (and compute new values for them). For example, the user-defined predicate `possibly-adjacent` in the rules in Figure 6 ensures that the steps are consecutive in some linearization of the plan.[9] For the plan in Figure 4 the extension of the `possibly-adjacent` predicate is: (0 4), (0 5), (4 5), (5 4), (4 1), (5 1), (1 2), (2 3), and (3 Goal).

---

8. Equality is denoted by sharing variables in the rule specification.

The user can easily add interpreted predicates by including a function definition that implements the predicate. During rule matching our algorithm passes arguments and calls such functions when appropriate. The current plan is passed as a default first argument to the interpreted predicates in order to provide a context for the computation of the predicate (but it can be ignored). Figure 11 show a skeleton for the (Lisp) implementation of the `possibly-adjacent` and `less-than` interpreted predicates.

```
(defun possibly-adjacent (plan node1 node2)      (defun less-than (plan n1 n2)
   (not (necessarily-not-adjacent                   (declare (ignore plan))
           node1                                     (when (and (numberp n1) (numberp n2))
           node2                                       (if (< n1 n2)
           ;; accesses the current plan                   '(nil) ;; true
           (plan-ordering plan)))                       nil))) ;; false
```

Figure 11: Sample Implementation of Interpreted Predicates

The consequent is composed of the `:replace` and `:with` fields. The `:replace` field specifies the subplan that is going to be removed from the plan, which is a subset of the steps and links identified in the antecedent. If a step is removed, all the links that refer to the step are also removed. The `:with` field specifies the replacement subplan. As we will see in Sections 3.1.2 and 3.1.3, the replacement subplan does not need to be completely specified. For example, the `:with` field of the `avoid-move-twice` rule of Figure 6 only specifies the addition of a `stack` step but not how this step is embedded into the plan. The links to the rest of the plan are automatically computed during the rewriting process.

### 3.1.2 PLAN-REWRITING RULES: FULL VERSUS PARTIAL SPECIFICATION

PbR gives the user total flexibility in defining rewriting rules. In this section we describe two approaches to guaranteeing that a rewriting rule specification preserves plan correctness, that is, produces a valid rewritten plan when applied to a valid plan.

In the *full-specification* approach the rule specifies *all* steps and links involved in a rewriting. The rule antecedent identifies all the anchoring points for the operators in the consequent, so that the embedding of the replacement subplan is unambiguous and results in a valid plan. The burden of proving the rule correct lies upon the user or an automated rule defining procedure (cf. Section 6). These kind of rules are the ones typically used in graph rewriting systems (Schürr, 1997).

In the *partial-specification* approach the rule defines the operators and links that constitute the *gist* of the plan transformation, but the rule does not prescribe the precise embedding of the replacement subplan. The burden of producing a valid plan lies upon the system. PbR takes advantage of the semantics of domain-independent planning to accept such a relaxed rule specification, fill in the details, and produce a valid rewritten plan. Moreover, the user is free to specify rules that may not necessarily be able to compute a rewriting for a plan that matches the antecedent because some necessary condition was not checked in the antecedent. That is, a partially-specified rule may be overgeneral. This may seem undesirable, but often a rule may cover more useful cases and be more naturally specified in this form. The rule may only fail for rarely occurring plans, so that the effort in defining and matching the complete specification may not be worthwhile. In any case, the plan-rewriting algorithm ensures that the application of a rewriting rule either generates a valid plan or fails to produce a plan (Theorem 1, Section 3.1.3).

---

9. The interpreted predicate `possibly-adjacent` makes the link expression in the antecedent of `avoid-move-twice` redundant. Unstack puts the block `?b1` on the table from where it is picked up by the stack operator, thus the causal link (?n1 (on ?b1 Table) ?n2) is already implied by the `:operators` and `:constraints` specification and could be removed from the rule specification.

As an example of these two approaches to rule specification, consider Figure 12 that shows the `avoid-move-twice-full` rule, a fully-specified version of the `avoid-move-twice` rule (of Figure 6, reprinted here for convenience). The `avoid-move-twice-full` rule is more complex and less natural to specify than `avoid-move-twice`. But, more importantly, `avoid-move-twice-full` is making more commitments than `avoid-move-twice`. In particular, `avoid-move-twice-full` fixes the producer of (`clear ?b1`) for `?n3` to be `?n4` when `?n7` is also known to be a valid candidate. In general, there are several alternative producers for a precondition of the replacement subplan, and consequently many possible embeddings. A different fully-specified rule is needed to capture each embedding. The number of rules grows exponentially as all permutations of the embeddings are enumerated. However, by using the partial-specification approach we can express a general plan transformation by a single natural rule.

```
(define-rule :name avoid-move-twice-full          (define-rule :name avoid-move-twice
  :if (:operators ((?n1 (unstack ?b1 ?b2))          :if (:operators
                  (?n2 (stack ?b1 ?b3 Table)))            ((?n1 (unstack ?b1 ?b2))
        :links ((?n4 (clear ?b1) ?n1)                      (?n2 (stack ?b1 ?b3 Table)))
                (?n5 (on ?b1 ?b2) ?n1)                :links (?n1 (on ?b1 Table) ?n2)
                (?n1 (clear ?b2) ?n6)                 :constraints
                (?n1 (on ?b1 Table) ?n2)                ((possibly-adjacent ?n1 ?n2)
                (?n7 (clear ?b1) ?n2)                    (:neq ?b2 ?b3)))
                (?n8 (clear ?b3) ?n2)            :replace (:operators (?n1 ?n2))
                (?n2 (on ?b1 ?b3) ?n9))          :with (:operators
        :constraints ((possibly-adjacent ?n1 ?n2)        (?n3 (stack ?b1 ?b3 ?b2))))
                (:neq ?b2 ?b3)))
  :replace (:operators (?n1 ?n2))
  :with (:operators ((?n3 (stack ?b1 ?b3 ?b2)))
        :links ((?n4 (clear ?b1) ?n3)
                (?n8 (clear ?b3) ?n3)
                (?n5 (on ?b1 ?b2) ?n3)
                (?n3 (on ?b1 ?b3) ?n9))))
```

Figure 12: Fully-specified versus Partially-specified Rewriting Rule

In summary, the main advantage of the full-specification rules is that the rewriting can be performed more efficiently because the embedding of the consequent is already specified. The disadvantages are that the number of rules to represent a generic plan transformation may be very large and the resulting rules quite lengthy; both of these problems may decrease the performance of the match algorithm. Also, the rule specification is error prone if written by the user. Conversely, the main advantage of the partial-specification rules is that a single rule can represent a complex plan transformation naturally and concisely. The rule can cover a large number of plan structures even if it may occasionally fail. Also, the partial specification rules are much easier to specify and understand by the users of the system. As we have seen, PbR provides a high degree of flexibility for defining plan-rewriting rules.

### 3.1.3 PLAN-REWRITING ALGORITHM

In this section, first we describe the basic plan-rewriting algorithm in PbR. Second, we prove this algorithm sound and discuss some formal properties of rewriting. Finally, we discuss a family of algorithms for plan rewriting depending on parameters such as the language for defining plan operators, the specification language for the rewriting rules, and the requirements of the search method.

The plan-rewriting algorithm is shown in Figure 13. The algorithm takes two inputs: a valid plan $P$, and a rewriting rule $R = (q_m, p_r, p_c)$ ($q_m$ is the antecedent query, $p_r$ is the replaced subplan, and $p_c$ is the replacement subplan). The output is a valid rewritten plan $P'$. The matching of the antecedent of the rewriting rule ($q_m$) determines if the rule is applicable and identifies the steps and links of interest (line 1). This matching can be seen as subgraph isomorphism between the antecedent subplan and the current plan (with the results then filtered by applying the :constraints). However, we take a different approach. PbR implements rule matching as conjunctive query evaluation. Our implementation keeps a relational representation of the steps and links in the current plan similar to the node and link specifications of the rewriting rules. For example, the database for the plan in Figure 4 contains one table for the unstack steps with schema (?n1 ?b1 ?b2) and tuples (4 C A) and (5 B D), another table for the causal links involving the clear condition with schema (?n1 ?n2 ?b) and tuples (0 1 C), (0 2 B), (0 2 C), (0 3 B), (0 4 C), (0 5 B), (4 3 A) and (5 1 D), and similar tables for the other operator and link types. The match process consists of interpreting the rule antecedent as a conjunctive query with interpreted predicates, and executing this query against the relational view of the plan structures. As a running example, we will analyze the application of the avoid-move-twice rule of Figure 6 to the plan in Figure 4. Matching the rule antecedent identifies steps 1 and 4. More precisely, considering the antecedent as a query, the result is the single tuple (4 C A 1 D) for the variables (?n1 ?b1 ?b2 ?n2 ?b3).

After choosing a match $\sigma_i$ to work on (line 3), the algorithm instantiates the subplan specified by the :replace field ($p_r$) according to such match (line 4) and removes the instantiated subplan $p_r^i$ from the original plan $P$ (line 5). All the edges incoming and emanating from nodes of the replaced subplan are also removed. The effects that the replaced plan $p_r^i$ was achieving for the remainder of the plan ($P - p_r^i$), the *UsefulEffects* of $p_r^i$, will now have to be achieved by the replacement subplan (or other steps of $P - p_r^i$). In order to facilitate this process, the *AddFlaws* procedure records these effects as open conditions.[10]

---

10. POCL planners operate by keeping track and repairing *flaws* found in a partial plan. Open conditions, operator threats, and resource threats are collectively called flaws (Penberthy & Weld, 1992). *AddFlaws(F,P)* adds the set of flaws $F$ to the plan structure $P$.

---

**procedure** *RewritePlan*

*Input*: a valid partial-order plan $P$
        a rewriting rule $R = (q_m, p_r, p_c)$, $Variables(p_r) \subseteq Variables(q_m)$

*Output*: a valid rewritten partial-order plan $P'$ (or failure)

1. $\Sigma := Match(q_m, P)$

   **Match** the rule antecedent $q_m$ (`:if` field) against $P$. The result is a set of substitutions $\Sigma = \{..., \sigma_i, ...\}$ for variables in $q_m$.

2. **If** $\Sigma = \emptyset$ **then return** failure

3. **Choose** a match $\sigma_i \in \Sigma$

4. $p_r^i := \sigma_i p_r$

   Instantiate the subplan to be removed $p_r$ (the `:replace` field) according to $\sigma_i$.

5. $P_r^i := AddFlaws(UsefulEffects(p_r^i), P - p_r^i)$

   **Remove** the instantiated subplan $p_r^i$ from the plan $P$ and add the UsefulEffects of $p_r^i$ as open conditions. The resulting plan $P_r^i$ is now incomplete.

6. $p_c^i := \sigma_i p_c$

   Instantiate the replacement subplan $p_c$ (the `:with` field) according to $\sigma_i$.

7. $P_c^i := AddFlaws(Preconditions(p_c^i) \cup FindThreats(P_r^i \cup p_c^i), P_r^i \cup p_c^i)$

   **Add** the instantiated replacement subplan $p_c^i$ to $P_r^i$. Find new threats and open conditions and add them as flaws. $P_c^i$ is potentially incomplete, having several flaws that need to be resolved.

8. $P' := rPOP(P_c^i)$

   **Complete** the plan using a partial-order causal-link planning algorithm (restricted to do only step reuse, but no step addition) in order to resolve threats and open conditions. $rPOP$ returns failure if no valid plan can be found.

9. **Return** $P'$

---

Figure 13: Plan-Rewriting Algorithm

The result is the partial plan $P_r^i$ (line 5). Continuing with our example, Figure 14(a) shows the plan resulting from removing steps 1 and 4 from the plan in Figure 4.

Finally, the algorithm embeds the instantiated replacement subplan $p_c^i$ into the remainder of the original plan (lines 6-9). If the rule is completely specified, the algorithm simply adds the (already instantiated) replacement subplan to the plan, and no further work is necessary. If the rule is partially specified, the algorithm computes the embeddings of the replacement subplan into the remainder of the original plan in three stages. First, the algorithm adds the instantiated steps and links of the replacement plan $p_c^i$ (line 6) into the current partial plan $P_r^i$ (line 7). Figure 14(b) shows the state of our example after $p_c^i$, the new `stack` step (6), has been incorporated into the plan. Note the open conditions (`clear A`) and `on(C D)`. Second, the *FindThreats* procedure computes the possible threats, both operator threats and resource conflicts, occurring in the $P_r^i \cup p_c^i$ partial plan (line 7); for example, the threat situation on the `clear(C)` proposition between step 6 and 2 in Figure 14(b). These threats and the preconditions of the replacement plan $p_c^i$ are recorded by *AddFlaws* resulting in the partial plan $P_c^i$. Finally, the algorithm completes the plan using $rPOP$, a partial-order causal-link planning procedure restricted to only reuse steps (i.e., no step addition) (line 8). *rPOP* allows us to support our expressive operator language and to have the flexibility for computing one or all embeddings. If only one rewriting is needed, *rPOP* stops at the first valid plan. Otherwise, it continues until exhausting all alternative ways of satisfying open preconditions and resolving conflicts, which produces all valid rewritings. In our running example, only one embedding is possible and the resulting plan is that of Figure 14(c), where the new `stack` step (6) produces (`clear A`) and `on(C D)`, its preconditions are satisfied, and the ordering (6 2) ensures that the plan is valid.

The rewriting algorithm in Figure 13 is sound in the sense that it produces a valid plan if the input is a valid plan, or it outputs failure if the input plan cannot be rewritten using the given rule. Since this elementary plan-rewriting step is sound, the sequence of rewritings performed during PbR's optimization search is also sound.

**Lemma 1 (Soundness of rPOP)** Partial-order causal-link (POCL) planning without step addition (*rPOP*) is sound.

*Proof:* In POCL planning, a precondition of a step of a plan is achieved either by inserting a new step $s_{new}$ or reusing a step $s_{reuse}$ already present in the current plan (the steps having an effect that unifies with the precondition). Forbidding step addition decreases the set of available steps that can be used to satisfy a precondition, but once a step is found rPOP proceeds as general POCL. Since, the POCL completion of a partial-plan is sound (Penberthy & Weld, 1992), *rPOP* is also sound. □

**Theorem 1 (Soundness of Plan Rewriting)** *RewritePlan* (Figure 13) produces a valid plan if the input $P$ is a valid plan, or outputs failure if the input plan cannot be rewritten using the given rewriting rule $R = (q_m, p_r, p_c)$.

*Proof:* Assume plan $P$ is a solution to a planning problem with goals $G$ and initial state $I$. In POCL planning, a plan is valid iff the preconditions of all steps are supported by causal links (the goals $G$ are the preconditions of the goal step, and the initial state conditions $I$ are the effects of the initial step), and no operator threatens any causal link (McAllester & Rosenblitt, 1991; Penberthy & Weld, 1992).

If rule $R$ does not match plan $P$, the algorithm trivially returns failure (line 2). Assuming there is a match $\sigma_i$, after removing from $P$ the steps and links specified in $p_r^i$ (including all links – causal and ordering – incoming and outgoing from steps of $p_r^i$), the only open conditions that exist in the resulting plan $P_r^i$ are those that $p_r^i$ was achieving (line 5). Adding the instantiated replacement subplan $p_c^i$ introduces more open conditions in the partial plan: the preconditions of the steps of $p_c^i$ (line 7). There are no other sources of open conditions in the algorithm.

Since plan $P$ is valid initially, the only (operator and/or resource) threats present in plan $P_c^i$ (line 7) are those caused by the removal of subplan $p_r^i$ (line 3) and the addition of subplan $p_c^i$ (line 7). The threats may occur between *any* operators and causal links of $P_r^i \cup p_c^i$ regardless whether the operator or causal link was initially in $P_r^i$ or in $p_c^i$. The threats in the combined plan $P_r^i \cup p_c^i$ can be effectively computed by finding the relative positions of its steps and comparing each causal link against the steps that may be ordered between the producer and the consumer of the condition in the causal link (*FindThreats*, line 7).

At this point, we have shown that we have a plan ($P_c^i$) with all the flaws (threats and open conditions) explicitly recorded (by *AddFlaws* in lines 5 and 7). Since $rPOP$ is sound (Lemma 1), we conclude that $rPOP$ will complete $P_c^i$ and output a valid plan $P'$, or output failure if the flaws in the plan cannot be repaired. □

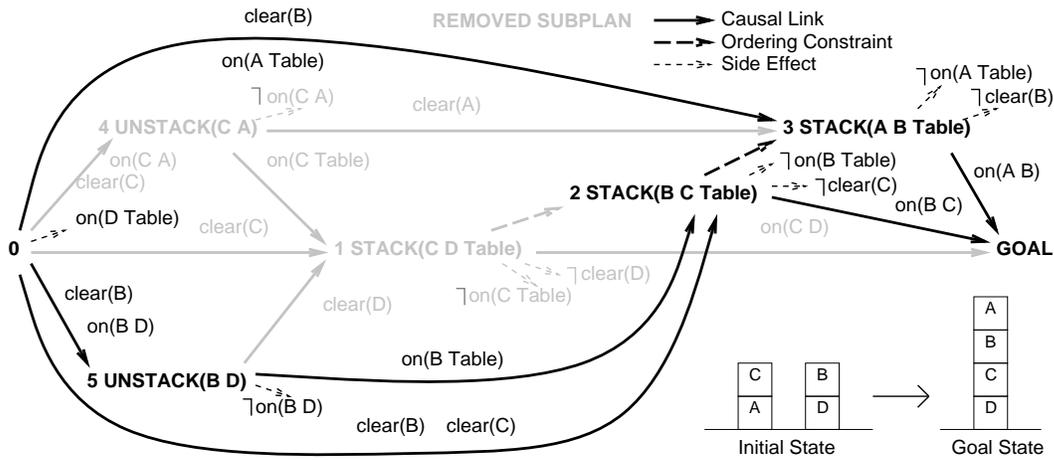

(a) Application of a Rewriting Rule: After Removing Subplan

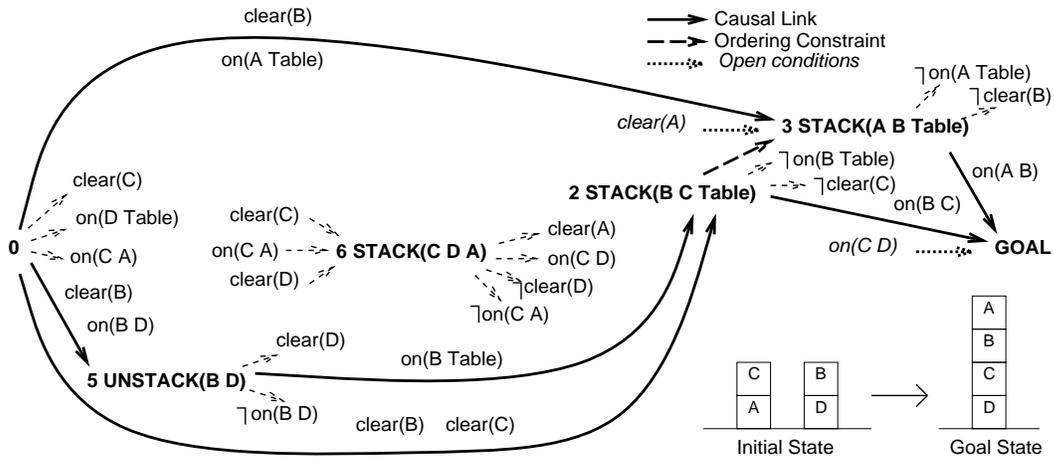

(b) Application of a Rewriting Rule: After Adding Replacement Subplan

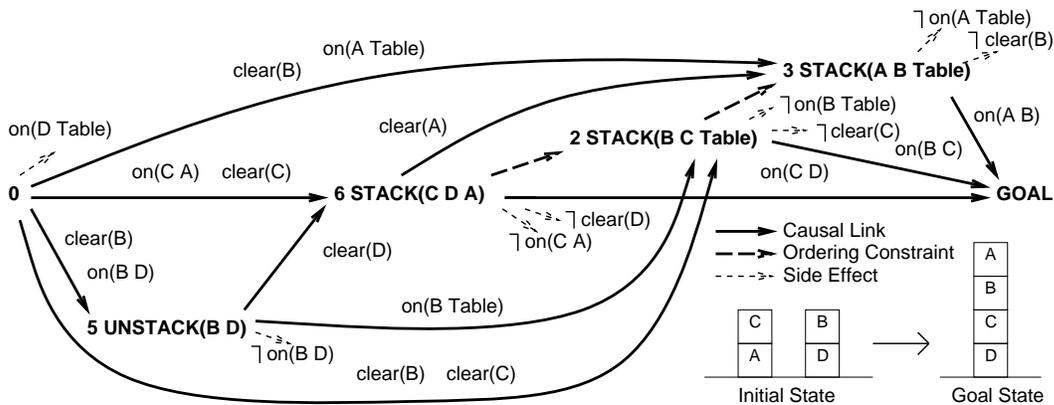

(c) Rewritten Plan

Figure 14: Plan Rewriting: Applying rule `avoid-move-twice` of Figure 6 to plan of Figure 4

**Corollary 1 (Soundness of PbR Search)** The optimization search of PbR is sound.

*Proof:* By induction. Assume an initial valid plan and a single step rewriting search. By Theorem 1, the output is either a valid rewritten plan or failure. If the output is failure, the search is trivially sound. Assume there is a valid plan $P_{n-1}$ after $n-1$ rewriting steps. According to Theorem 1, applying a single rewriting rule to plan $P_{n-1}$ produces a valid plan $P_n$ or failure. Thus, an arbitrary number of rewritings produces a valid plan (or no plan), so PbR's search is sound. □

Although *RewritePlan* is sound, it may certainly produce plans that do not have the minimal number of steps when faced with arbitrary rules. For example, imagine that the consequent of a rewriting rule specified two identical steps s1 and s2 (both having as only effects e1 and e2) and that the only flaws in $P_c^i$ were exactly the open conditions e1 and e2. Then, a sound but non step-minimal plan would be using s1 to satisfy e1 and using s2 to satisfy e2 (although each step by itself could satisfy both open conditions). PbR does not discard this plan because we do not make any restriction on the types of acceptable cost functions. If we had a cost function that took the robustness of the plan into account, a plan with both steps may be desirable.

We cannot guarantee that PbR's optimization search is complete in the sense that the optimal plan would be found. PbR uses local search and it is well known that, in general, local search cannot be complete. Even if PbR exhaustively explores the space of plan rewritings induced by a given initial plan and a set of rewriting rules, we still cannot prove that all solution plans will be reached. This is a property of the initial plan generator, the set of rewriting rules, and the semantics of the planning domain. The rewriting rules of PbR play a similar role as traditional declarative search control where the completeness of the search may be traded for efficiency. Perhaps using techniques for inferring invariants in a planning domain (Gerevini & Schubert, 1998; Fox & Long, 1998; Rintanen, 2000) or proving convergence of term and graph rewriting systems (Baader & Nipkow, 1998), conditions for completeness of a plan-rewriting search in a given planning domain could be obtained.

The design of a plan-rewriting algorithm depends on several parameters: the language of the operators, the language of the rewriting rules, the choice of full-specification or partial-specification rewriting rules, and the need for all rewritings or one rewriting as required by the search method.

The *language of the operators* affects the way in which the initial and rewritten plans are constructed. Our framework supports the expressive operator definition language described in Section 2.1. We provide support for this language by using standard techniques for causal link establishment and threat checking like those in Sage (Knoblock, 1995) and UCPOP (Penberthy & Weld, 1992).

The *language of the antecedents of the rewriting rules* affects the efficiency of matching. Our system implements the conjunctive query language that was described in Section 3.1.1. However, our system could easily accommodate a more expressive query language for the rule antecedent, such as a relationally complete language (i.e., conjunction, disjunction, and safe negation) (Abiteboul, Hull, & Vianu, 1995), or a recursive language such as datalog with stratified negation, without significantly increasing the computational complexity of the approach in an important way, as we discuss in Section 3.1.4.

The choice of *fully versus partially specified rewriting rules* affects the way in which the replacement plan is embedded into the current plan. If the rule is completely specified, the embedding is already specified in the rule consequent, and the replacement subplan is simply added to the current plan. If the rule is partially specified, our algorithm can compute all the valid embeddings.

The choice of *one versus all rewritings* affects both the antecedent matching and the embedding of rule consequent. The rule matches can be computed either all at the same time, as in bottom-up evaluation of logic databases, or one-at-a-time as in Prolog, depending on whether the search strategy requires one or all rewritings. If the rule is fully-specified only one embedding per match is possible. But, if the rule is partially-specified multiple embeddings may result from a single match. If the search strategy only requires one rewriting, it must also provide a mechanism for choosing which rule is applied, which match is computed, and which embedding is generated (*rPOP* can stop at the first embedding or compute all embeddings). Our implemented rewriting algorithm has a modular design to support different combinations of these choices.

### 3.1.4 COMPLEXITY OF PLAN REWRITING

The complexity of plan rewriting in PbR originates from two sources: matching the rule antecedent against the plan, and computing the embeddings of the replacement plan. In order to analyze the complexity of matching plan-rewriting rules, we introduce the following database-theoretic definitions of complexity (Abiteboul et al., 1995):

**Data Complexity:** complexity of evaluating a *fixed query* for variable database inputs.

**Expression Complexity:** complexity of evaluating, on a *fixed database instance*, the queries specifiable in a given query language.

Data complexity measures the complexity with respect to the size of the database. Expression complexity measures the complexity with respect to the size of the queries (taken from a given language). In our case, the database is the steps and links of the plan and the queries are the antecedents of the plan-rewriting rules.

Formally, the language of the rule antecedents described in Section 3.1.1 is conjunctive queries with interpreted predicates. The worst-case combined data and expression complexity of conjunctive queries is exponential (Abiteboul et al., 1995). That is, if the size of the query (rule antecedent) and the size of the database (plan) grow simultaneously, there is little hope of matching efficiently. Fortunately, relationally-complete languages have a *data* complexity contained in Logarithmic Space, which is, in turn, contained in Polynomial Time (Abiteboul et al., 1995). Thus our conjunctive query language has at most this complexity. This is a very encouraging result that shows that the cost of evaluating a fixed query grows very slowly as the database size increases. For PbR this means that matching the antecedent of the rules is not strongly affected by the size of the plans. Moreover, in our experience useful rule antecedents are not very large and contain many constant labels (at least, the node and edge predicate names) that help to reduce the size of the intermediate results and improve the efficiency of matching. This result also indicates that we could extend the language of the antecedent to be relationally complete without affecting significantly the performance of the system.[11] Another possible extension is to use datalog with stratified negation, which also has polynomial time data complexity. Graph-theoretic properties of our plans could be easily described in datalog. For example, the `possibly-adjacent` interpreted predicate of Figure 7 could be described declaratively as a datalog program instead of a piece of code. In summary, rule match for moderately sized rules, even for quite expressive languages and large plans, remains tractable and can be made efficient using production match (Forgy, 1982) and query optimization techniques (Sellis, 1988).

---

11. Figure 32 in Section 6 proposes an example of a rule with a relationally-complete antecedent using an appropriate syntax.

The second source of complexity is computing the embeddings of the replacement plan given in the consequent of a plan-rewriting rule. By the definition of full-specification rules, the embedding is completely specified in the rule itself. Thus, it suffices simply to remove the undesired subplan and directly add the replacement subplan. This is linear in the size of the consequent.

For partial-specification rules, computing all the embeddings of the replacement subplan can be exponential in the size of the plan in the worst case. However, this occurs only in pathological cases. For example, consider the plan in Figure 15(a) in which we are going to compute the embeddings of step **x** into the remainder of the plan in order to satisfy the open precondition *g0*. Step **x** has no preconditions and has two effects ¬b and **g0**. Each step in the plan has proposition **b** as an effect. Therefore, the new step **x** conflicts with every step in the plan (**1** to **n**) and has to be ordered with respect to these steps. Unfortunately, there are an exponential number of orderings. In effect, the orderings imposed by adding the step **x** correspond to *all* the partitions of the set of steps (**1** to **n**) into two sets: one ordered before **x** and one after. Figure 15(b) shows one of the possible orderings. If the subplan we were embedding contained several steps that contained similar conflicts the problem would be compounded. Even deciding if a single embedding exists is NP-hard. For example, if we add two additional effects ¬a and ¬g1 to operator **x**, there is no valid embedding. In the worst case (solving first the flaws induced by the conflicts on proposition **b**) we have to explore an exponential number of positions for step **x** in the plan, all of which end up in failure. Nevertheless, given the quasi-decomposability of useful planning domains we expect the number of conflicts to be relatively small. Also most of the useful rewriting rules specify replacement subplans that are small compared with the plan they are embedding into. Our experience indicates that plan rewriting with partial-specification rules can be performed efficiently as shown by the results of Section 4.

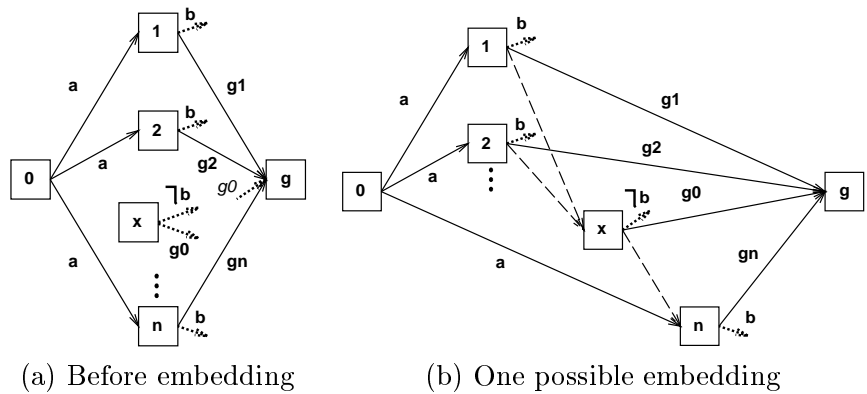

(a) Before embedding      (b) One possible embedding

Figure 15: Exponential Embeddings

### 3.1.5 A Taxonomy of Plan-Rewriting Rules

In order to guide the user in defining plan-rewriting rules for a domain or to help in designing algorithms that may automatically deduce the rules from the domain specification (see Section 6), it is helpful to know what kinds of rules are useful. We have identified the following general types of transformation rules:

**Reorder:** These are rules based on algebraic properties of the operators, such as commutative, associative and distributive laws. For example, the commutative rule that reorders two operators that need the same resource in Figure 10, or the `join-swap` rule in Figure 29 that combines the commutative and associative properties of the relational algebra.

**Collapse:** These are rules that replace a subplan by a smaller subplan. For example, when several operators can be replaced by one, as in the `remote-join-eval` rule in Figure 29. This rule replaces two remote retrievals at the same information source and a local join operation by a single remote join operation, when the remote source has the capability of performing joins. An example of the application of this rule to a query plan is shown in Figure 30. Other examples are the Blocks World rules in Figure 6 that replace an `unstack` and a `stack` operators either by an equivalent single `stack` operator or the empty plan.

**Expand:** These are rules that replace a subplan by a bigger subplan. Although this may appear counter-intuitive initially, it is easy to imagine a situation in which an expensive operator can be replaced by a set of operators that are cheaper as a whole. An interesting case is when some of these operators are already present in the plan and can be synergistically reused. We did not find this rule type in the domains analyzed so far, but Bäckström (1994a) presents a framework in which adding actions improves the quality of the plans. His quality metric is the plan execution time, similarly to the manufacturing domain of Section 4.1. Figure 16 shows an example of a planning domain where adding actions improves quality (from Bäckström, 1994a). In this example, removing the link between `Bm` and `C1` and inserting a new action `A'` shortens significantly the time to execute the plan.

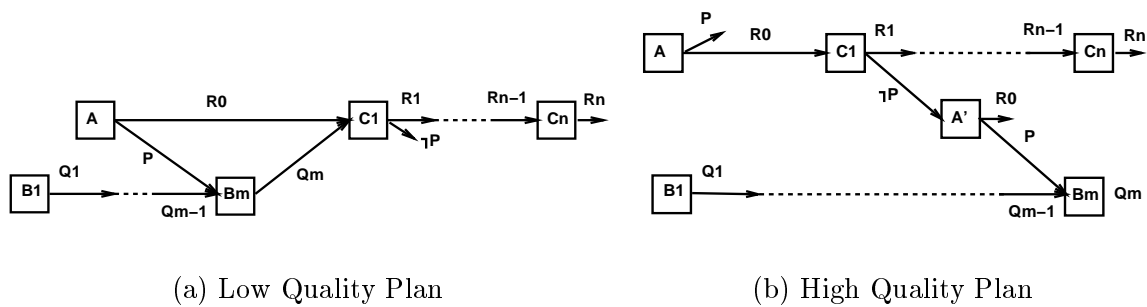

(a) Low Quality Plan          (b) High Quality Plan

Figure 16: Adding Actions Can Improve Quality

**Parallelize:** These are rules that replace a subplan with an equivalent alternative subplan that requires fewer ordering constraints. A typical case is when there are redundant or alternative resources that the operators can use. For example, the rule `punch-by-drill-press` in Figure 7. Another example is the rule that Figure 16 suggests that could be seen as a combination of the expand and parallelize types.

## 3.2 Selection of Next Plan: Search Strategies

Although the space of rewritings can be explored systematically, the Planning by Rewriting framework is better suited to the local search techniques typical of combinatorial optimization algorithms. The characteristics of the planning domain, the initial plan generator, and the rewriting rules determine which local search method performs best. First, we discuss how the initial plan generator affects the choice of local search methods. Second, we consider the impact of the rewriting rules. Third, we discuss the role of domain knowledge in the search process. Finally, we describe how several local search methods work in PbR.

An important difference between PbR and traditional combinatorial algorithms is the generation of feasible solutions. Usually, in combinatorial optimization problems there exists an effective procedure to generate *all* feasible solutions (e.g., the permutations of a schedule). Thus, even if the local search graph is disconnected, by choosing an appropriate initial solution generator (e.g., random) we could fall in a component of the graph that contains the global optimum. In PbR we cannot assume such powerful initial plan generators. Even in optimization domains, which have efficient initial plan generators, we may not have guarantees on the coverage of the solution space they provide. Therefore, the optimal plan may not be reachable by applying the rewriting rules when starting from the initial plans available from the generator. Nevertheless, for many domains an initial plan generator that provides a good sample of the solution space is sufficient for multiple-restart search methods to escape from low-quality local minima and provide high-quality solutions.

The plan-rewriting rules define the neighborhood function, which may be *exact* (cf. Section 2.3) or not. For example, in the query planning domain we can define a set of rules that completely generate the space of solution plans (because of the properties of the relational algebra). In other domains it may be hard to prove that we have an exact set of rules. Both the limitations on initial plan generation and the plan-rewriting rules affect the possibility of theoretically reaching the global optimum. This is not surprising since many problems, regardless of whether they are cast as planning or in other formalisms, do not have converging local search algorithms (e.g., Papadimitriou & Steiglitz, 1977). Nevertheless, in practice, good local optima can still be obtained for many domains.

Many local search methods, such as first and best improvement, simulated annealing, tabu search, or variable-depth search, can be applied straightforwardly to PbR. In our experiments in Section 4 we have used first and best improvement, which have performed well. Next, we describe some details of the application of these two methods in PbR. In Section 6, we discuss our ideas for using variable-depth plan rewriting.

*First improvement* generates the rewritings incrementally and selects the first plan of better cost than the current one. In order to implement this method efficiently we can use a tuple-at-a-time evaluation of the rule antecedent, similarly to the behavior of Prolog. Then, for that rule instantiation, generate one embedding, test the cost of the resulting plan, and if it is not better that the current plan, repeat. We have the choice of generating another embedding of the same rule instantiation, generate another instantiation of the same rule, or generate a match for a different rule.

*Best improvement* generates the complete set of rewritten plans and selects the best. This method requires computing all matches and all embeddings for each match. All the matches can be obtained by evaluating the rule antecedent as a set-at-a-time database query. As we discussed in Section 3.1.4 such query evaluation can be quite efficient. In our experience, computing the plan embeddings was usually more expensive than computing the rule matches.

In Planning by Rewriting the choice of the initial plan generator, the rewriting rules, and the search methods is intertwined. Once the initial plan generator is fixed, it determines the shape of the plans that would have to be modified by the rewriting rules, then according to this neighborhood, the most appropriate search mechanism can be chosen. PbR has a modular design to facilitate experimentation with different initial plan generators, sets of rewriting rules, and search strategies.

## 3.3 Plan Quality

In most practical planning domains the quality of the plans is crucial. This is one of the motivations for the Planning by Rewriting approach. In PbR the user defines the measure of plan quality most appropriate for the application domain. This quality metric could range from a simple domain-independent cost metric, such as the number of steps, to more complex domain-specific ones. For example, in the query planning domain the measure of plan quality usually is an estimation of the query execution cost based on the size of the database relations, the data manipulation operations involved in answering a query, and the cost of network transfer. In a decentralized environment, the cost metric may involve actual monetary costs if some of the information sources require payments. In the job-shop scheduling domain some simple cost functions are the schedule length (that is, the parallel time to finish all pieces), or the sum of the times to finish each piece. A more sophisticated manufacturing domain may include a variety of concerns such as the cost, reliability, and precision of each operator/process, the costs of resources and materials used by the operators, the utilization of the machines, etc. The reader will find more detailed examples of quality metrics in these domains in Sections 4.1 and 4.4.

A significant advantage of PbR is that the complete plan is available to assess its quality. In generative planners the complete plan is not available until the search for a solution is completed, so usually only very simple plan quality metrics, such as the number of steps, can be used. Some work does incorporate quality concerns into generative planners (Estlin & Mooney, 1997; Borrajo & Veloso, 1997; Pérez, 1996). These systems automatically learn search control rules to improve both the efficiency of planning and the quality of the resulting plans. In PbR the rewriting rules can be seen as "post facto" optimization search control. As opposed to guiding the search of a generative planner towards high-quality solutions based only on the information available in partial plans, PbR improves the quality of complete solution plans without any restriction on the types of quality metrics. Moreover, if the plan cost is not additive, a plan refinement strategy is impractical since it may need to exhaustively explore the search space to find the optimal plan. An example of non-additive cost function appears in the UNIX planning domain (Etzioni & Weld, 1994) where a plan to transfer files between two machines may be cheaper if the files are compressed initially (and uncompressed after arrival). That is, the plan that includes the compression (and the necessary uncompression) operations is more cost effective, but a plan refinement search would not naturally lead to it. By using complete plans, PbR can accurately assess arbitrary measures of quality.

## 3.4 Initial Plan Generation

Fast initial plan generation is domain-specific in nature. It requires the user to specify an efficient mechanism to compute the initial solution plan. In general, generating an initial plan may be as hard as generating the optimal plan. However, the crucial intuition behind planning algorithms is that most practical problems are quasi-decomposable (Simon, 1969), that is, that the interactions among parts of the problems are limited. If interactions in a problem are pervasive, such as in the 8-puzzle, the operator-based representation and the algorithms of classical planning are of little use. They would behave as any other search based problem solver. Fortunately, many practical problems are indeed quasi-decomposable. This same intuition also suggests that finding initial plan generators for planning problems may not be as hard as it appears, because the system can solve the subproblems independently, and then combine them in the simplest way, for example, concatenating the solutions sequentially. Moreover, in many circumstances the problems may be easily transformed into a state that minimizes the interactions and solving the problem from this state is much easier. For example, in the Blocks World the state in which all blocks are on the table minimizes the interactions. It is simple to design an algorithm that solves any Blocks World problem passing through such intermediate state. Using these methods an initial plan generator may produce suboptimal initial plans but at a reasonable planning cost.

These ideas for constructing initial plan generators can be embodied in two general ways, which are both implemented in our system. The first one is to bootstrap on the results of a general purpose planning algorithm with a strong search control bias. The second one is to provide the user convenient high-level facilities in which to describe plan construction algorithms programmatically.

### 3.4.1 BIASED GENERATIVE PLANNERS

There are a variety of ways in which to control the search of a generic planner. Some planners accept search control rules, others accept heuristic functions, and some have built-in search control. We present examples of these techniques.

A very general way of efficiently constructing plans is to use a domain-independent generative planner that accepts search control rules. For example, Prodigy (Carbonell, Knoblock, & Minton, 1991), UCPOP (Penberthy & Weld, 1992) and Sage (Knoblock, 1995) are such planners. By setting the type of search and providing a strong bias by means of the search control rules, the planner can quickly generate a valid, although possibly suboptimal, initial plan. For example, in the manufacturing domain of (Minton, 1988a), analyzed in detail in Section 4.1, depth-first search and a goal selection heuristic based on abstraction hierarchies (Knoblock, 1994a) quickly generates a feasible plan, but often the quality of this plan, which is defined as the time required to manufacture all objects, is suboptimal.

TLPlan (Bacchus & Kabanza, 1995, 2000) is an efficient forward-chaining planner that uses search control expressed in temporal logic. Because in forward chaining the complete state is available, much more refined domain control knowledge can be specified. The preferred search strategy used by TLPlan is depth-first search, so although it finds plans efficiently, the plans may be of low quality. Note that because it is a generative planner that explores partial sequences of steps, it cannot use sophisticated quality measures.

HSP (Bonet, Loerincs, & Geffner, 1997; Bonet & Geffner, 1999) is a forward search planner that performs a variation of heuristic search applied to classical AI planning. The built-in heuristic function is a relaxed version of the planning problem: it computes the number of required steps to reach the goal disregarding negated effects in the operators. Such metric can be computed efficiently. Despite its simplicity and that the heuristic is not admissible, it scales surprisingly well for many domains. Because the plans are generated according to the fixed heuristic function, the planner cannot incorporate a quality metric.

These types of planners are quite efficient in practice although they often produce suboptimal plans. They are excellent candidates to generate the initial plans that will be subsequently optimized by PbR.

### 3.4.2 Facilitating Algorithmic Plan Construction

For many domains, simple domain-dependent approximation algorithms will provide good initial plans. For example, in the query planning domain, the system can easily generate initial query evaluation plans by randomly (or greedily) parsing the given query. In the Blocks World it is also straightforward to generate a solution in linear time using the naive algorithm: put all blocks on the table and build the desired towers from the bottom up. This algorithm produces plans of length no worse than twice the optimal, which makes it already a good approximation algorithm. However, the interest in the Blocks World has traditionally been on optimal solutions, which is an NP-hard problem (Gupta & Nau, 1992).

Our system facilitates the creation of these initial plans by freeing the user from specifying the detailed graph structure of a plan. The user only needs to specify an algorithm that produces a sequence of instantiated actions, that is, action names and the ground parameters that each action takes.[12] For example, the (user-defined) naive algorithm for the Blocks World domain described above applied to the problem in Figure 4 produces the sequence: `unstack(C A)`, `unstack(B D)`, `stack(C D Table)`, `stack(B C Table)`, and `stack(A B Table)`. Then, the system automatically converts this sequence of actions into a fully detailed partial-order plan using the operator specification of the domain. The resulting plan conforms to the internal data structures that PbR uses. This process includes creating nodes that are fully detailed operators with preconditions and effects, and adding edges that represent all the necessary causal links and ordering constraints. In our Blocks World example the resulting plan is that of Figure 4.

The algorithm that transforms the user-defined sequence of actions into a partial-order plan is presented in Figure 17. The algorithm first constructs the causal structure of the plan (lines 2 to 6) and then adds the necessary ordering links to avoid threats (lines 7 to 10). The user only needs to specify action names and the corresponding instantiated action parameters. Our algorithm consults the operator specification to find the preconditions and effects, instantiate them, construct the causal links, and check for operator threats. Operator threats are always resolved in favor of the ordering given by the user in the input plan. The reason is that the input plan may be overconstrained by the total order, but it is assumed valid. Therefore, by processing each step last to first, only the orderings that indeed avoid threats are included in the partial-order plan.

---

12. The algorithm also accepts extra ordering constraints in addition to the sequence if they are available from the initial plan generator.

---

**procedure** TO2PO
*Input*: a valid total-order plan $(a_1, ..., a_n)$
*Output*: an equivalent partial-order plan
1. **for** $i := n$ to 1
2.     **for** $p \in Preconditions(a_i)$
3.         **choose** $k < i$ such that
4.             1. $p \in PositiveEffects(a_k) \wedge$
5.             2. $\not\exists l$ such that $k < l < i \wedge p \in NegativeEffects(a_l)$
6.         add order $a_k \prec a_i$
7.     **for** $p \in NegativeEffects(a_i)$
8.         **for** $j := (i - 1)$ to 1
9.             **if** $p \in Preconditions(a_j)$
10.                 **then** add order $a_j \prec a_i$
11. **return** $((a_1, ..., a_n), \prec)$

---

Figure 17: Algorithm for Converting Total-order to Partial-order Plans

Our algorithm is an extension of the greedy algorithm presented by Veloso, Perez, & Carbonell (1990). Our algorithm explores non-deterministically all the producers of a proposition (line 3), as opposed to taking the latest producer in the sequence as in their algorithm.[13] That is, if our algorithm is explored exhaustively, it produces all partially-ordered causal structures consistent with the input sequence. Our generalization stems from the criticism by Bäckström (1994b) to the algorithm by Veloso et al. (1990) and our desire of being able to produce alternative initial plans.

The problem of transforming a sequence of steps into a least constrained plan is analyzed by Bäckström (1994b) under several natural definitions of optimality. Under his definitions of least-constrained plan and shortest parallel execution the problem is NP-hard. Bäckström shows that Veloso's algorithm, although polynomial, does not conform to any of these natural definitions. Because our algorithm is not greedy, it does not suffer from the drawbacks pointed out by Bäckström. Moreover, for our purposes we do not need optimal initial plans. The space of partial orders will be explored during the rewriting process.

Regardless of the method for producing initial plans, generators that provide multiple plans are preferable. The different initial plans are used in conjunction with multiple restart search techniques in order to escape from low-quality local minima.

## 4. Empirical Results

In this section we show the broad applicability of Planning by Rewriting by analyzing four domains with different characteristics: a process manufacturing domain (Minton, 1988b), a transportation logistics domain, the Blocks World domain that we used in the examples throughout the paper, and a domain for distributed query planning.

---

13. To implement their algorithm it is enough to replace line 3 in Figure 17 with:
    find max $k < i$ such that

## 4.1 Manufacturing Process Planning

The task in the manufacturing process planning domain is to find a plan to manufacture a set of parts. We implemented a PbR translation of the domain specification in (Minton, 1988b). This domain contains a variety of machines, such as a lathe, punch, spray painter, welder, etc, for a total of ten machining operations. The operator specification is shown in Figures 18 and 19. The features of each part are described by a set of predicates, such as `temperature`, `painted`, `has-hole`, etc. These features are changed by the operators. Other predicates in the state, such as `has-clamp`, `is-drillable`, etc, are set in the initial state of each problem.

As an example of the behavior of an operator, consider the `polish` operator in Figure 18. It requires the part to manufacture to be cold and that the polisher has a clamp to secure the part to the machine. The effect of applying this operator is to leave the surface of the part polished. Some attributes of a part, such as `surface-condition`, are single-valued, but others, like `has-hole`, are multivalued. Note how the `drill-press` and the `punch` operators in Figure 18 do not prevent several `has-hole` conditions from being asserted on the same part. Other interesting operators are `weld` and `bolt`. These operators join two parts in a particular orientation to form a new part. No further operations can be performed on the separate parts once they have been joined.

The measure of plan cost is the schedule length, the (parallel) time to manufacture *all* parts. In this domain all of the machining operations are assumed to take unit time. The machines and the objects (parts) are modeled as resources in order to enforce that only one part can be placed on a machine at a time and that a machine can only operate on a single part at a time (except `bolt` and `weld` which operate on two parts simultaneously).

We have already shown some of the types of rewriting rules for this domain in Figures 7 and 10. The set of rules that we used for our experiments is shown in Figure 20. The top eight rules are quite straightforward once one becomes familiar with this domain. The two top rules explore the space of alternative orderings originated by resource conflicts. The `machine-swap` rule allows the system to explore the possible orderings of operations that require the same machine. This rule finds two consecutive operations on the same machine and swaps their order. Similarly, the rule `object-swap` allows the system to explore the orderings of the operations on the same object. These two rules use the interpreted predicate `adjacent-in-critical-path` to focus the attention on the steps that contribute to our cost function. `Adjacent-in-critical-path` checks if two steps are consecutive along one of the critical paths of a schedule. A critical path is a sequence of steps that take the longest time to accomplish. In other words, a critical path is one of the sequences of steps that determine the schedule length.

The next six rules exchange operators that are equivalent with respect to achieving some effects. Rules `IP-by-SP` and `SP-by-IP` propose the exchange of `immersion-paint` and `spray-paint` operators. By examining the operator definitions in Figure 19, it can be readily noticed that both operators change the value of the `painted` predicate. Similarly, `PU-by-DP` and `DP-by-PU` exchange `drill-press` and `punch` operators, which produce the `has-hole` predicate. Finally, `roll-by-lathe` and `lathe-by-roll` exchange `roll` and `lathe` operators as they both can make parts cylindrical. To focus the search on the most promising

```
(define (operator POLISH)                      (define (operator GRIND)
 :parameters (?x)                               :parameters (?x)
 :resources ((machine POLISHER) (is-object ?x)) :resources ((machine GRINDER) (is-object ?x))
 :precondition (:and (is-object ?x)             :precondition (is-object ?x)
                     (temperature ?x COLD)      :effect
                     (has-clamp POLISHER))       (:and (:forall (?color)
 :effect                                               (:not (painted ?x ?color)))
  (:and (:forall (?surf)                             (:forall (?surf)
          (:when (:neq ?surf POLISHED)                 (:when (:neq ?surf SMOOTH)
             (:not (surface-condition ?x ?surf)))         (:not (surface-condition ?x ?surf))))
        (surface-condition ?x POLISHED)))           (surface-condition ?x SMOOTH)))

(define (operator LATHE)                       (define (operator ROLL)
 :parameters (?x)                               :parameters (?x)
 :resources ((machine LATHE) (is-object ?x))    :resources ((machine ROLLER) (is-object ?x))
 :precondition (is-object ?x)                   :precondition (is-object ?x)
 :effect                                        :effect
  (:and (:forall (?color)                        (:and (:forall (?color)
          (:not (painted ?x ?color)))                  (:not (painted ?x ?color)))
        (:forall (?shape)                            (:forall (?shape)
          (:when (:neq ?shape CYLINDRICAL)             (:when (:neq ?shape CYLINDRICAL)
             (:not (shape ?x ?shape))))                  (:not (shape ?x ?shape))))
        (:forall (?surf)                             (:forall (?temp)
          (:when (:neq ?surf ROUGH)                     (:when (:neq ?temp HOT)
             (:not (surface-condition ?x ?surf))))        (:not (temperature ?x ?temp))))
        (surface-condition ?x ROUGH)                 (:forall (?surf)
        (shape ?x CYLINDRICAL)))                       (:not (surface-condition ?x ?surf)))
                                                     (:forall (?width ?orientation)
                                                       (:not (has-hole ?x ?width ?orientation)))
                                                     (temperature ?x HOT)
                                                     (shape ?x CYLINDRICAL)))

(define (operator DRILL-PRESS)                 (define (operator PUNCH)
 :parameters (?x ?width ?orientation)           :parameters (?x ?width ?orientation)
 :resources ((machine DRILL-PRESS)              :resources ((machine PUNCH) (is-object ?x))
             (is-object ?x))                    :precondition
 :precondition                                   (:and (is-object ?x)
  (:and (is-object ?x)                                 (has-clamp PUNCH)
        (have-bit ?width)                             (is-punchable ?x ?width ?orientation))
        (is-drillable ?x ?orientation))        :effect
 :effect (has-hole ?x ?width ?orientation))     (:and (:forall (?surf)
                                                       (:when (:neq ?surf ROUGH)
                                                          (:not (surface-condition ?x ?surf))))
                                                     (surface-condition ?x ROUGH)
                                                     (has-hole ?x ?width ?orientation)))
```

Figure 18: Operators for Manufacturing Process Planning (I)

exchanges these rules only match operators in the critical path (by means of the interpreted predicate `in-critical-path`).

The six bottom rules in Figure 20 are more sophisticated. The `lathe+SP-by-SP` rule takes care of an undesirable effect of the simple depth-first search used by our initial plan generator. In this domain, in order to spray paint a part, the part must have a regular shape. Being cylindrical is a regular shape, therefore the initial planner may decide to make the part cylindrical by lathing it in order to paint it! However, this may not be necessary as the part may already have a regular shape (for example, it could be rectangular, which is also a regular shape). Thus, the `lathe+SP-by-SP` substitutes the pair `spray-paint` and `lathe` by a single `spray-paint` operation. The supporting `regular-shapes` interpreted predicate

```
(define (operator IMMERSION-PAINT)          (define (operator SPRAY-PAINT)
 :parameters (?x ?color)                     :parameters (?x ?color ?shape)
 :resources ((machine IMMERSION-PAINTER)     :resources ((machine SPRAY-PAINTER)
            (is-object ?x))                              (is-object ?x))
 :precondition                               :precondition (:and (is-object ?x)
   (:and (is-object ?x)                                       (sprayable ?color)
         (have-paint-for-immersion ?color))                  (temperature ?x COLD)
 :effect (painted ?x ?color))                               (regular-shape ?shape)
                                                            (shape ?x ?shape)
                                                            (has-clamp SPRAY-PAINTER))
                                             :effect (painted ?x ?color))

(define (operator WELD)                      (define (operator BOLT)
 :parameters (?x ?y ?new-obj ?orient)        :parameters (?x ?y ?new-obj ?orient ?width)
 :resources ((machine WELDER)                :resources ((machine BOLTER) (is-object ?y))
            (is-object ?x) (is-object ?y))   :precondition
 :precondition                                 (:and (is-object ?x) (is-object ?y)
   (:and (is-object ?x) (is-object ?y)              (composite-object ?new-obj ?orient ?x ?y)
         (composite-object ?new-obj ?orient ?x ?y) (has-hole ?x ?width ?orient)
         (can-be-welded ?x ?y ?orient))            (has-hole ?y ?width ?orient)
 :effect (:and (temperature ?new-obj HOT)          (bolt-width ?width)
               (joined ?x ?y ?orient)              (can-be-bolted ?x ?y ?orient))
               (:not (is-object ?x))         :effect (:and (:not (is-object ?x))
               (:not (is-object ?y))))                    (:not (is-object ?y))
                                                          (joined ?x ?y ?orient)))
```

Figure 19: Operators for Manufacturing Process Planning (II)

just enumerates which are the regular shapes. These rules are partially specified and are not guaranteed to always produce a rewriting. Nevertheless, they are often successful in producing plans of lower cost.

The remaining rules explore bolting two parts using bolts of different size if fewer operations may be needed for the plan. We developed these rules by analyzing differences in the quality of the optimal plans and the rewritten plans. For example, consider the `both-providers-diff-bolt` rule. This rule states that if the parts to be bolted already have compatible holes in them, it is better to reuse those operators that produced the holes. The initial plan generator may have drilled (or punched) holes whose only purpose was to bolt the parts. However, the goal of the problem may already require some holes to be performed on the parts to be joined. Reusing the available holes produces a more economical plan. The rules `has-hole-x-diff-bolt-add-PU`, `has-hole-x-diff-bolt-add-DP`, `has-hole-y-diff-bolt-add-PU`, and `has-hole-y-diff-bolt-add-DP` address the cases in which only one of the holes can be reused, and thus an additional `punch` or `drill-press` operation needs to be added.

As an illustration of the rewriting process in the manufacturing domain, consider Figure 21. The plan at the top of the figure is the result of a simple initial plan generator that solves each part independently and concatenates the corresponding subplans. Although such plan is generated efficiently, it is of poor quality. It requires six time-steps to manufacture all parts. The figure shows the application of two rewriting rules, `machine-swap` and `IP-by-SP`, that improve the quality of this plan. The operators matched by the rule antecedent are shown in *italics*. The operators introduced in the rule consequent are shown in **bold**. First, the `machine-swap` rule reorders the punching operations on parts **A** and **B**. This

```
(define-rule :name machine-swap                    (define-rule :name object-swap
  :if (:operators ((?n1 (machine ?x) :resource)      :if (:operators ((?n1 (is-object ?x) :resource)
                   (?n2 (machine ?x) :resource))                       (?n2 (is-object ?x) :resource))
       :links ((?n1 :threat ?n2))                         :links ((?n1 :threat ?n2))
       :constraints                                       :constraints
         (adjacent-in-critical-path ?n1 ?n2))                (adjacent-in-critical-path ?n1 ?n2))
  :replace (:links (?n1 ?n2))                        :replace (:links (?n1 ?n2))
  :with (:links (?n2 ?n1)))                          :with (:links (?n2 ?n1)))

(define-rule :name IP-by-SP                         (define-rule :name SP-by-IP
  :if (:operators (?n1 (immersion-paint ?x ?c))      :if (:operators (?n1 (spray-paint ?x ?c ?s))
       :constraints ((regular-shapes ?s)                  :constraints ((in-critical-path ?n1)))
                     (in-critical-path ?n1)))        :replace (:operators (?n1))
  :replace (:operators (?n1))                        :with (:operators (?n2 (immersion-paint ?x ?c))))
  :with (:operators (?n2 (spray-paint ?x ?c ?s))))

(define-rule :name PU-by-DP                         (define-rule :name DP-by-PU
  :if (:operators (?n1 (punch ?x ?w ?o))             :if (:operators ((?n1 (drill-press ?x ?w ?o)))
       :constraints ((in-critical-path ?n1)))             :constraints ((in-critical-path ?n1)))
  :replace (:operators (?n1))                        :replace (:operators (?n1))
  :with (:operators (?n2 (drill-press ?x ?w ?o))))   :with (:operators (?n2 (punch ?x ?w ?o))))

(define-rule :name roll-by-lathe                    (define-rule :name lathe-by-roll
  :if (:operators ((?n1 (roll ?x)))                  :if (:operators ((?n1 (lathe ?x)))
       :constraints ((in-critical-path ?n1)))             :constraints ((in-critical-path ?n1)))
  :replace (:operators (?n1))                        :replace (:operators (?n1))
  :with (:operators (?n2 (lathe ?x))))               :with (:operators (?n2 (roll ?x))))

(define-rule :name lathe+SP-by-SP                   (define-rule :name both-providers-diff-bolt
  :if (:operators                                    :if (:operators ((?n3 (bolt ?x ?y ?z ?o ?w1)))
        ((?n1 (lathe ?x))                                 :links ((?n1 (has-hole ?x ?w1 ?o) ?n3)
         (?n2 (spray-paint ?x ?color ?shape1)))                   (?n2 (has-hole ?y ?w1 ?o) ?n3)
       :constraints ((regular-shapes ?shape2)))               (?n4 (has-hole ?x ?w2 ?o) ?n5)
  :replace (:operators (?n1 ?n2))                            (?n6 (has-hole ?y ?w2 ?o) ?n7))
  :with (:operators                                       :constraints ((:neq ?w1 ?w2)))
         ((?n3 (spray-paint ?x ?color ?shape2)))))  :replace (:operators (?n1 ?n2 ?n3))
                                                     :with (:operators ((?n8 (bolt ?x ?y ?z ?o ?w2)))
                                                           :links ((?n4 (has-hole ?x ?w2 ?o) ?n8)
                                                                   (?n6 (has-hole ?y ?w2 ?o) ?n8))))

(define-rule :name has-hole-x-diff-bolt-add-PU      (define-rule :name has-hole-x-diff-bolt-add-DP
  :if (:operators ((?n3 (bolt ?x ?y ?z ?o ?w1)))     :if (:operators ((?n3 (bolt ?x ?y ?z ?o ?w1)))
       :links ((?n1 (has-hole ?x ?w1 ?o) ?n3)             :links ((?n1 (has-hole ?x ?w1 ?o) ?n3)
               (?n2 (has-hole ?y ?w1 ?o) ?n3)                     (?n2 (has-hole ?y ?w1 ?o) ?n3)
               (?n4 (has-hole ?x ?w2 ?o) ?n5))                    (?n4 (has-hole ?x ?w2 ?o) ?n5))
       :constraints ((:neq ?w1 ?w2)))                     :constraints ((:neq ?w1 ?w2)))
  :replace (:operators (?n1 ?n2 ?n3))                :replace (:operators (?n1 ?n2 ?n3))
  :with (:operators ((?n8 (bolt ?x ?y ?z ?o ?w2))    :with (:operators ((?n8 (bolt ?x ?y ?z ?o ?w2))
                     (?n6 (punch ?x ?w2 ?o)))                          (?n6 (drill-press ?y ?w2 ?o)))
       :links ((?n4 (has-hole ?x ?w2 ?o) ?n8)             :links ((?n4 (has-hole ?x ?w2 ?o) ?n8)
               (?n6 (has-hole ?y ?w2 ?o) ?n8))))                   (?n6 (has-hole ?y ?w2 ?o) ?n8))))

(define-rule :name has-hole-y-diff-bolt-add-PU      (define-rule :name has-hole-y-diff-bolt-add-DP
  :if (:operators ((?n3 (bolt ?x ?y ?z ?o ?w1)))     :if (:operators ((?n3 (bolt ?x ?y ?z ?o ?w1)))
       :links ((?n1 (has-hole ?x ?w1 ?o) ?n3)             :links ((?n1 (has-hole ?x ?w1 ?o) ?n3)
               (?n2 (has-hole ?y ?w1 ?o) ?n3)                     (?n2 (has-hole ?y ?w1 ?o) ?n3)
               (?n6 (has-hole ?y ?w2 ?o) ?n7))                    (?n6 (has-hole ?y ?w2 ?o) ?n7))
       :constraints ((:neq ?w1 ?w2)))                     :constraints ((:neq ?w1 ?w2)))
  :replace (:operators (?n1 ?n2 ?n3))                :replace (:operators (?n1 ?n2 ?n3))
  :with (:operators ((?n8 (bolt ?x ?y ?z ?o ?w2))    :with (:operators ((?n8 (bolt ?x ?y ?z ?o ?w2))
                     (?n4 (punch ?x ?w2 ?o)))                          (?n4 (drill-press ?x ?w2 ?o)))
       :links ((?n4 (has-hole ?x ?w2 ?o) ?n8)             :links ((?n4 (has-hole ?x ?w2 ?o) ?n8)
               (?n6 (has-hole ?y ?w2 ?o) ?n8))))                   (?n6 (has-hole ?y ?w2 ?o) ?n8))))
```

Figure 20: Rewriting Rules for Manufacturing Process Planning

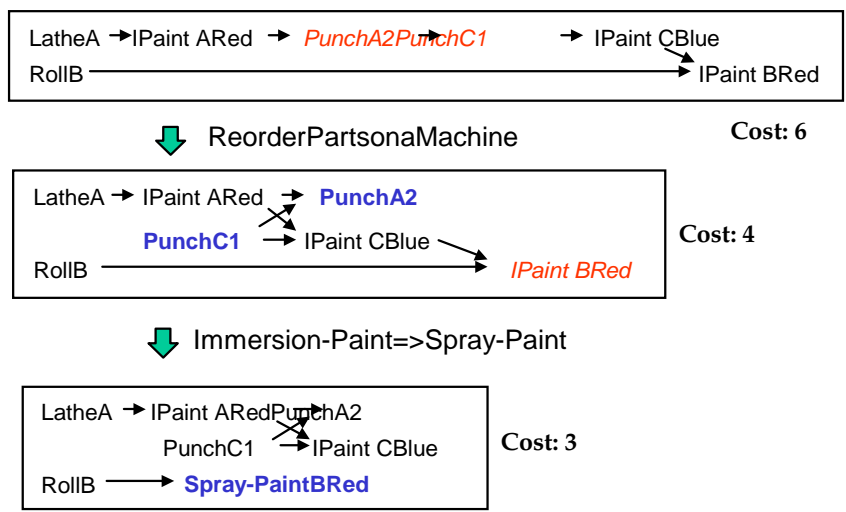

Figure 21: Rewriting in the Manufacturing Domain

breaks the long critical path that resulted from the simple concatenation of their respective subplans. The schedule length improves from six to four time-steps. Still, the three parts A, B, and C use the same painting operation (immersion-paint). As the immersion-painter can only process one piece at a time, the three operations must be done serially. Fortunately, in our domain there is another painting operation: spray-paint. The IP-by-SP rule takes advantage of this fact and substitutes an immersion-paint operation on part B by a spray-paint operation. This further parallelizes the plan obtaining a schedule length of three time-steps, which is the optimal for this plan.

We compare four planners (IPP, Initial, and two configurations of PbR):

**IPP:** This is one of the most efficient domain-independent planners (Koehler, Nebel, Hoffman, & Dimopoulos, 1997) in the planning competition held at the Fourth International Conference on Artificial Intelligence Planning Systems (AIPS-98). IPP is an optimized reimplementation and extension of Graphplan (Blum & Furst, 1995, 1997). IPP produces shortest parallel plans. For our manufacturing domain, this is exactly the schedule length, the cost function that we are optimizing.

**Initial:** The initial plan generator uses a divide-and-conquer heuristic in order to generate plans as fast as possible. First, it produces subplans for each part and for the joined goals independently. These subplans are generated by Sage using a depth-first search without any regard to plan cost. Then, it concatenates the subsequences of actions and merges them using the facilities of Section 3.4.2.

**PbR:** We present results for two configurations of PbR, which we will refer to as PbR-100 and PbR-300. Both configurations use a first improvement gradient search strategy with random walk on the cost plateaus. The rewriting rules used are those of Figure 20. For each problem PbR starts its search from the plan generated by Initial. The two configurations differ only on how many total plateau plans are allowed. PbR-100 allows considering up to 100 plans that do not improve the cost without terminating the search. Similarly, PbR-300 allows 300 plateau plans. Note that the limit is across all plateaus encountered during the search for a problem, not for each plateau.

We tested each of the four systems on 200 problems, for machining 10 parts, ranging from 5 to 50 goals. The goals are distributed randomly over the 10 parts. So, for the 50-goal problems, there is an average of 5 goals per part. The results are shown in Figure 22. In these graphs each data point is the average of 20 problems for each given number of goals. There were 10 provably unsolvable problems. Initial and PbR solved all 200 problems (or proved them unsolvable). IPP solved 65 problems in total: all problems at 5 and 10 goals, 19 at 15 goals, and 6 at 20 goals. IPP could not solve any problem with more than 20 goals under the 1000 CPU seconds time limit.

Figure 22(a) shows the average time on the solvable problems for each problem set for the four planners. Figure 22(b) shows the average schedule length for the problems solved by *all* the planners, that is, over the 65 problems solved by IPP up to 20 goals. The fastest planner is Initial, but it produces plans with a cost of about twice the optimal. IPP produces the optimal plans, but it cannot solve problems of more than 20 goals. The two configurations of PbR scale much better than IPP solving all problems and producing good quality plans. PbR-300 matches the optimal cost of the IPP plans, except in one problem (the reason for the difference is interesting and we explain it below). The faster PbR-100 also stays very close to the optimal (less than 2.5% average cost difference).

Figure 22(c) shows the average schedule length for the problems solved by *each* of the planners for the 50 goal range. The PbR configurations scale gracefully across this range improving considerably the cost of the plans generated by Initial. The additional exploration of PbR-300 allows it to improve the plans even further. The reason for the difference between PbR and IPP at the 20-goal complexity level is because the cost results for IPP are only for the 6 problems that it could solve, while the results for PbR and Initial are the average of all of the 20 problems (as shown in Figure 22(b), PbR matches the cost of these 6 optimal plans produced by IPP).

Figure 22(d) shows the average number of operators in the plans for the problems solved by *all* three planners (up to 20 goals). Figure 22(e) shows the average number of operators in the plans for the problems solved by *each* planner across the whole range of 50 problems. The plans generated by Initial use about 2-3 additional operators. Both PbR and IPP produce plans that require fewer steps. Interestingly, IPP sometimes produces plans that use more operations than PbR. IPP produces the shortest parallel plan, but not the one with the minimum number of steps. In particular, we observed that some of the IPP plans suffer from the same problem as Initial. IPP would also lathe a part in order to paint it, but as opposed to Initial it would only do so if it did not affect the optimal schedule length. Surprisingly, adding such additional steps in this domain may improve the schedule length, albeit in fairly rare situations. This was the case in the only problem in which IPP produced a better schedule than PbR-300. We could have introduced a rewriting rule that substituted an `immersion-paint` operator by both a `lathe` and `spray-paint` operators for such cases. However, such rule is of very low utility (in the sense of Minton, 1988b). It expands the rewriting search space, adds to the cost of match, and during the random search provides some benefit very rarely.

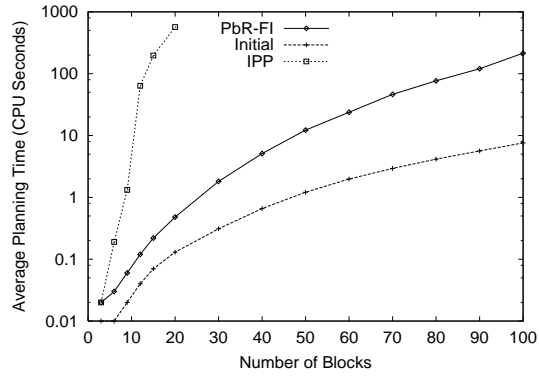

(a) Average Planning Time

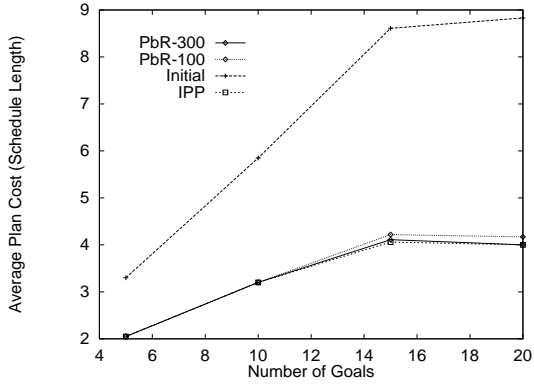

(b) Average Plan Cost
(Problems Solved by All)

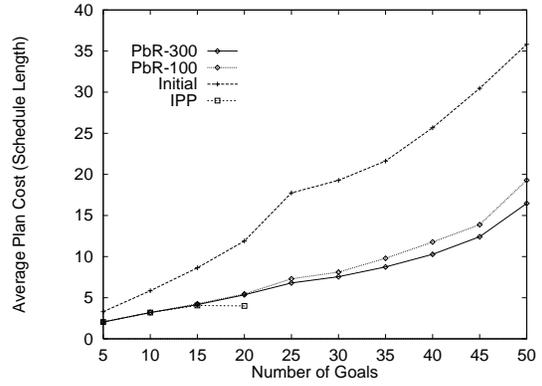

(c) Average Plan Cost
(Problems Solved by Each)

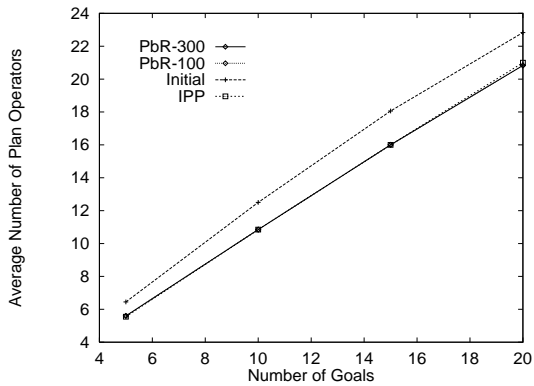

(d) Number of Plan Operators
(Problems Solved by All)

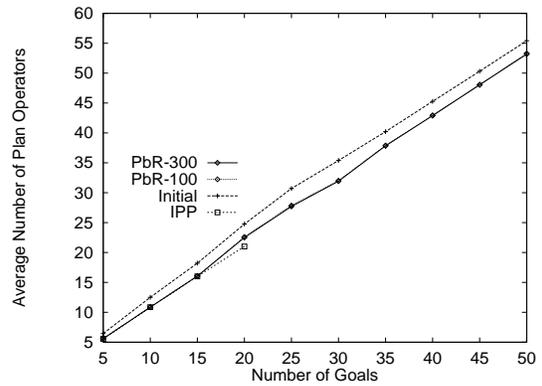

(e) Number of Plan Operators
(Problems Solved by Each)

Figure 22: Experimental Results: Manufacturing Process Planning

This experiment illustrates the flexibility of PbR in specifying complex rules for a planning domain. The results show the benefits of finding a suboptimal initial plan quickly and then efficiently transforming it to improve its quality.

## 4.2 Logistics

The task in the logistics domain is to transport several packages from their initial location to their desired destinations. We used a version of the `logistics-strips` planning domain of the AIPS98 planning competition which we restricted to using only trucks but not planes.[14] The domain is shown in Figure 23. A package is transported from one location to another by loading it into a truck, driving the truck to the destination, and unloading the truck. A truck can load any number of packages. The cost function is the (parallel) time to deliver all packages (measured as the number of operators in the critical path of a plan).

```
(define (operator LOAD-TRUCK)              (define (operator UNLOAD-TRUCK)
 :parameters (?obj ?truck ?loc)            :parameters (?obj ?truck ?loc)
 :precondition                             :precondition
  (:and (obj ?obj) (truck ?truck) (location ?loc)   (:and (obj ?obj) (truck ?truck) (location ?loc)
        (at ?truck ?loc) (at ?obj ?loc))          (at ?truck ?loc) (in ?obj ?truck))
 :effect (:and (:not (at ?obj ?loc))       :effect (:and (:not (in ?obj ?truck))
               (in ?obj ?truck)))                        (at ?obj ?loc)))

(define (operator DRIVE-TRUCK)
 :parameters (?truck ?loc-from ?loc-to ?city)
 :precondition (:and (truck ?truck) (location ?loc-from) (location ?loc-to) (city ?city)
                     (at ?truck ?loc-from) (in-city ?loc-from ?city) (in-city ?loc-to ?city))
 :effect (:and (:not (at ?truck ?loc-from)) (at ?truck ?loc-to)))
```

Figure 23: Operators for Logistics

We compare three planners on this domain:

**IPP:** IPP (Koehler et al., 1997) produces optimal plans in this domain.

**Initial:** The initial plan generator picks a distinguished location and delivers packages one by one starting and returning to the distinguished location. For example, assume that truck t1 is at the distinguished location l1, and package p1 must be delivered from location l2 to location l3. The plan would be: `drive-truck(t1 l1 l2 c)`, `load-truck(p1 t1 l2)`, `drive-truck(t1 l2 l3 c)`, `unload-truck(p1 t1 l3)`, `drive-truck(t1 l3 l1 c)`. The initial plan generator would keep producing these circular trips for the remaining packages. Although this algorithm is very efficient it produces plans of very low quality.

**PbR:** PbR starts from the plan produced by Initial and uses the plan rewriting rules shown in Figure 24 to optimize plan quality. The `loop` rule states that driving to a location and returning back immediately after is useless. The fact that the operators must be adjacent is important because it implies that no intervening `load` or `unload` was performed. In the same vein, the `triangle` rule states that it is better to drive directly between two locations than through a third point if no other operation is performed at such point. The

---

14. In the logistics domain of AIPS98, the problems of moving packages by plane among different cities and by truck among different locations in a city are isomorphic, so we focused on only one of them to better analyze how the rewriting rules can be learned (Ambite, Knoblock, & Minton, 2000).

**load-earlier** rule captures the situation in which a package is not loaded in the truck the first time that the package's location is visited. This occurs when the initial planner was concerned with a trip for another package. The **unload-later** rule captures the dual case. PbR applies a first improvement search strategy with only one run (no restarts).

```
(define-rule :name loop
  :if (:operators
       ((?n1 (drive-truck ?t ?l1 ?l2 ?c))
        (?n2 (drive-truck ?t ?l2 ?l1 ?c)))
       :links ((?n1 ?n2))
       :constraints
       ((adjacent-in-critical-path ?n1 ?n2)))
  :replace (:operators (?n1 ?n2))
  :with NIL)

(define-rule :name load-earlier
  :if (:operators
       ((?n1 (drive-truck ?t ?l1 ?l2 ?c))
        (?n2 (drive-truck ?t ?l3 ?l2 ?c))
        (?n3 (load-truck ?p ?t ?l2)))
       :links ((?n2 ?n3))
       :constraints
       ((adjacent-in-critical-path ?n2 ?n3)
        (before ?n1 ?n2)))
  :replace (:operators (?n3))
  :with (:operators ((?n4 (load-truck ?p ?t ?l2)))
         :links ((?n1 ?n4))))
```

```
(define-rule :name triangle
  :if (:operators
       ((?n1 (drive-truck ?t ?l1 ?l2 ?c))
        (?n2 (drive-truck ?t ?l2 ?l3 ?c)))
       :links ((?n1 ?n2))
       :constraints
       ((adjacent-in-critical-path ?n1 ?n2)))
  :replace (:operators (?n1 ?n2))
  :with (:operators
         ((?n3 (drive-truck ?t ?l1 ?l3 ?c)))))

(define-rule :name unload-later
  :if (:operators
       ((?n1 (drive-truck ?t ?l1 ?l2 ?c))
        (?n2 (unload-truck ?p ?t ?l2))
        (?n3 (drive-truck ?t ?l3 ?l2 ?c)))
       :links ((?n1 ?n2))
       :constraints
       ((adjacent-in-critical-path ?n1 ?n2)
        (before ?n2 ?n3)))
  :replace (:operators (?n2))
  :with (:operators ((?n4 (unload-truck ?p ?t ?l2)))
         :links ((?n3 ?n4))))
```

Figure 24: Logistics Rewriting Rules

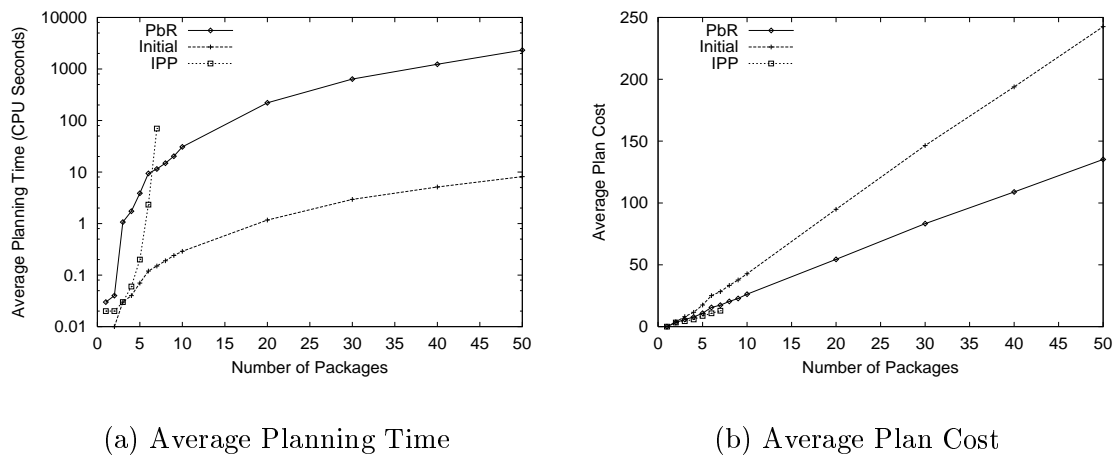

(a) Average Planning Time          (b) Average Plan Cost

Figure 25: Experimental Results: Logistics, Scaling the Number of Packages

We compared the performance of IPP, Initial, and PbR on a set of logistics problems involving up to 50 packages. Each problem instance has the same number of packages, locations, and goals. There was a single truck and a single city. The performance results are shown in Figure 25. In these graphs each data point is the average of 20 problems for each given number of packages. All the problems were satisfiable. IPP could only solve problems up to 7 packages (it also solved 10 out of 20 for 8 packages, and 1 out of 20 for 9 packages, but these are not shown in the figure). Figure 25(a) shows the average planning time. Figure 25(b) shows the average cost for the 50 packages range. The results are similar to the previous experiment. Initial is efficient but highly suboptimal. PbR is able to considerably improve the cost of these plans and approach the optimal.

## 4.3 Blocks World

We implemented a classical Blocks World domain with the two operators in Figure 2. This domain has two actions: `stack` that puts one block on top of another, and, `unstack` that places a block on the table to start a new tower. Plan quality in this domain is simply the number of steps. Optimal planning in this domain is NP-hard (Gupta & Nau, 1992). However, it is trivial to generate a correct, but suboptimal, plan in linear time using the naive algorithm: put all blocks on the table and build the desired towers from the bottom up. We compare three planners on this domain:

**IPP:** In this experiment we used the GAM goal ordering heuristic (Koehler, 1998; Koehler & Hoffmann, 2000) that had been tested in Blocks World problems with good scaling results.

**Initial:** This planner is a programmatic implementation of the naive algorithm using the facilities introduced in Section 3.4.2.

**PbR:** This configuration of PbR starts from the plan produced by Initial and uses the two plan-rewriting rules shown in Figure 6 to optimize plan quality. PbR applies a first improvement strategy with only one run (no restarts).

We generated random Blocks World problems scaling the number of blocks. The problem set consists of 25 random problems at 3, 6, 9, 12, 15, 20, 30, 40, 50, 60, 70, 80, 90, and 100 blocks for a total of 350 problems. The problems may have multiple towers in the initial state and in the goal state.

Figure 26(a) shows the average planning time of the 25 problems for each block quantity. IPP cannot solve problems with more than 20 blocks within the time limit of 1000 CPU seconds. The problem solving behavior of IPP was interesting. IPP either solved a given problem very fast or it timed out. For example, it was able to solve 11 out of the 25 20-block problems under 100 seconds, but it timed out at 1000 seconds for the remaining 14 problems. This seems to be the typical behavior of complete search algorithms (Gomes, Selman, & Kautz, 1998). The local search of PbR allows it to scale much better and solve all the problems.

Figure 26(b) shows the average plan cost as the number of blocks increases. PbR improves considerably the quality of the initial plans. The optimal quality is only known for very small problems, where PbR approximates it, but does not achieve it (we ran Sage for problems of less than 9 blocks). For larger plans we do not know the optimal cost. However, Slaney & Thiébaux (1996) performed an extensive experimental analysis of Blocks World planning using a domain like ours. In their comparison among different approximation algorithms they found that our initial plan generator (unstack-stack) achieves empirically a quality around 1.22 the optimal for the range of problem sizes we have analyzed (Figure 7 in Slaney & Thiébaux, 1996). The value of our average initial plans divided by 1.22 suggests the quality of the optimal plans. The quality achieved by PbR is comparable with that value. In fact it is slightly better which may be due to the relatively small number of problems tested (25 per block size) or to skew in our random problem generator. Interestingly the plans found by IPP are actually of low quality. This is due to the fact that IPP produces shortest parallel plans. That means that the plans can be constructed in the fewest time steps, but IPP may introduce more actions in each time step than are required.

In summary, the experiments in this and the previous sections show that across a variety of domains PbR scales to large problems while still producing high-quality plans.

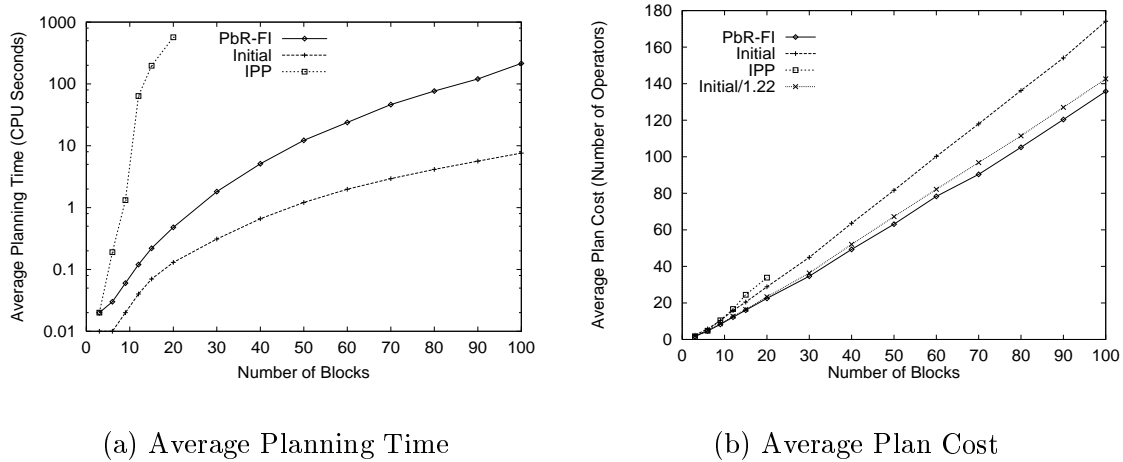

(a) Average Planning Time                     (b) Average Plan Cost

Figure 26: Experimental Results: Blocks World, Scaling the Number of Blocks

## 4.4 Query Planning

Query Planning is a problem of considerable practical importance. It is central to traditional database and mediator systems. In this section we present some results in distributed query planning to highlight the use of PbR in a domain with a complex cost function. A detailed description of query planning, including a novel query processing algorithm for mediators based on PbR, and a more extensive experimental analysis appear in (Ambite & Knoblock, 2000; Ambite, 1998).

Query planning involves generating a plan that efficiently computes a user query from the relevant information sources. This plan is composed of data retrieval actions at distributed information sources and data manipulation operations, such as those of the relational algebra: join, selection, union, etc. The specification of the operators for query planning and the encoding of information goals that we are using was first introduced by Knoblock (1996). A sample information goal is shown in Figure 27. This goal asks to send to the output device of the mediator all the names of airports in Tunisia. Two sample operators are shown in Figure 28. The `retrieve` operator executes a query at a remote information source and transports the data to the mediator, provided that the source is in operation (`source-available`) and that the source is capable of processing the query (`source-acceptable-query`). The `join` operator takes two subqueries, which are available locally at the mediator, and combines them using some conditions to produce the joined query.

```
(available sims (retrieve (?ap_name)
                    (:and (airport ?aport)
                          (country-name ?aport "Tunisia")
                          (port-name ?aport ?ap_name))))
```

Figure 27: Sample Information Goal

```
(define (operator retrieve)
   :parameters (?source ?query)
   :resources ((processor ?source))
   :precondition (:and (source-available ?source)
                       (source-acceptable-query ?query ?source))
   :effect (available sims ?query))

(define (operator join)
   :parameters (?join-conds ?query ?query-a ?query-b)
   :precondition (:and (available sims ?query-a)
                       (available sims ?query-b)
                       (join-query ?query ?join-conds ?query-a ?query-b))
   :effect (available sims ?query))
```

Figure 28: Some Query Planning Operators

The quality of a distributed query plan is an estimation of its execution cost, which is a function of the size of intermediate results, the cost of performing data manipulation operations, and the transmission through the network of the intermediate results from the remote sources to the mediator. Our system estimates the plan cost based on statistics obtained from the source relations, such as the number of tuples in a relation, the number of distinct values for each attribute, and the maximum and minimum values for numeric attributes (Silberschatz, Korth, & Sudarshan, 1997, chapter 12). The sources accessed, and the type and ordering of the data processing operations are critical to the plan cost.

The rewriting rules are derived from properties of the distributed environment and the relational algebra.[15] The first set of rules rely on the fact that, in a distributed environment, it is generally more efficient to execute a group of operations together at a remote information source than to transmit the data over the network and execute the operations at the local system. As an example consider the `Remote-Join-Eval` rule in Figure 29 (shown here in the PbR syntax, it was shown algebraically in Figure 1). This rule specifies that if in a plan there exist two retrieval operations at the same remote database whose results are consequently joined and the remote source is capable of performing joins, the system can rewrite the plan into one that contains a single retrieve operation that pushes the join to the remote database.

The second class of rules are derived from the commutative, associative, and distributive properties of the operators of the relational algebra. For example, the `Join-Swap` rule of Figure 29 (cf. Figure 1) specifies that two consecutive joins operators can be reordered and allows the planner to explore the space of join trees. Since in our query planning

---

15. In mediators, rules that address the resolution of the semantic heterogeneity are also necessary. See (Ambite & Knoblock, 2000; Ambite, 1998) for details.

```
(define-rule :name remote-join-eval        (define-rule :name join-swap
 :if (:operators                            :if (:operators
      ((?n1 (retrieve ?query1 ?source))          ((?n1 (join ?q1 ?jc1 ?sq1a ?sq1b))
       (?n2 (retrieve ?query2 ?source))           (?n2 (join ?q2 ?jc2 ?sq2a ?sq2b)))
       (?n3 (join ?query ?jc ?query1 ?query2)))   :links (?n2 ?n1)
      :constraints                               :constraints
       ((capability ?source 'join)))             (join-swappable
 :replace (:operators (?n1 ?n2 ?n3))               ?q1 ?jc1 ?sq1a ?sq1b      ;; in
 :with (:operators                                 ?q2 ?jc2 ?sq2a ?sq2b      ;; in
       ((?n4 (retrieve ?query ?source))))          ?q3 ?jc3 ?sq3a ?sq3b      ;; out
                                                   ?q4 ?jc4 ?sq4a ?sq4b))    ;; out
                                            :replace (:operators (?n1 ?n2))
                                            :with (:operators
                                                  ((?n3 (join ?q3 ?jc3 ?sq3a ?sq3b))
                                                   (?n4 (join ?q4 ?jc4 ?sq4a ?sq4b))
                                                   :links (?n4 ?n3)))
```

Figure 29: Some Query Planning Rewriting Rules

domain queries are expressed as complex terms (Knoblock, 1996), the PbR rules use the interpreted predicates in the `:constraints` field to manipulate such query expressions. For example, the `join-swappable` predicate checks if the queries in the two join operators can be exchanged and computes the new subqueries.

Figure 30 shows an example of the local search through the space of query plan rewritings in a simple distributed domain that describes a company. The figure shows alternative query evaluation plans for a conjunctive query that asks for the names of employees, their salaries, and the projects they are working on. The three relations requested in the query (`Employees`, `Payroll`, and `Project`) are distributed among two databases (one at the company's headquarters – `HQ-db` – and another at a branch – `Branch-db`). Assume that the leftmost plan is the initial plan. This plan first retrieves the `Employee` relation at the `HQ-db` and the `Project` relation at the `Branch-db`, and then it joins these two tables on the employee `name`. Finally, the plan retrieves the `Payroll` relation from the `HQ-db` and joins it on `ssn` with the result of the previous join. Although a valid plan, this initial plan is suboptimal. Applying the `join-swap` rule to this initial plan generates two rewritings. One of them involves a cross-product, which is a very expensive operation, so the system, following a gradient descent search strategy, prefers the other plan. Now the system applies the `remote-join-eval` rule and generates a new rewritten plan that evaluates the join between the employee and project tables remotely at the headquarters database. This final plan is of much better quality.

We compare the planning efficiency and plan quality of four query planners:

**Sage:** This is the original query planner (Knoblock, 1995, 1996) for the SIMS mediator, which performs a best-first search with a heuristic commonly used in query optimization that explores only the space of left join trees. Sage is a refinement planner (Kambhampati, Knoblock, & Yang, 1995) that generates optimal left-tree query plans.

**DP:** This is our implementation of a dynamic-programming bottom-up enumeration of query plans (Ono & Lohman, 1990) to find the optimal plan. Since in our distributed domain subqueries can execute in parallel and the cost function reflects such preference, our DP algorithm considers bushy join trees. However, to improve its planning time, DP applies the heuristic of avoiding cross-products during join enumeration. Thus, in some rare cases DP may not produce the optimal plan.

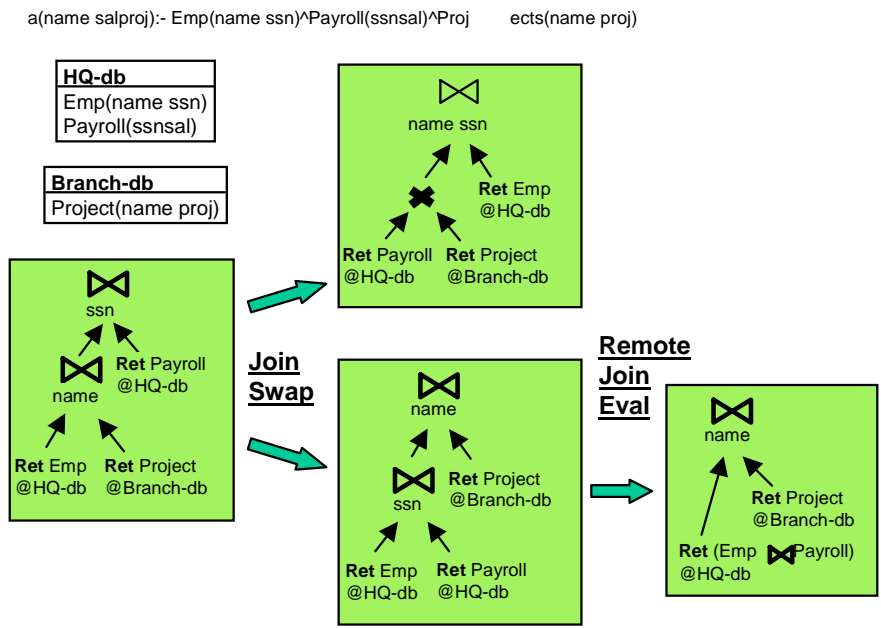

Figure 30: Rewriting in Query Planning

**Initial:** This is the initial plan generator for PbR. It generates query plans according to a random depth-first search parse of the query. The only non-random choice is that it places selections as soon as they can be executed. It is the fastest planner but may produce very low quality plans.

**PbR:** We used the `Remote-Join-Eval` and `Join-Swap` rules defined in Figure 29. These two rules are sufficient to optimize the queries in the test set. We tested two gradient-descent search strategies for PbR: first improvement with four random restarts (PbR-FI), and steepest descent with three random restarts (PbR-SD).

In this experiment we compare the behavior of Sage, DP, Initial, PbR-FI, and PbR-SD in a distributed query planning domain as the size of the queries increases. We generated a synthetic domain for the SIMS mediator and defined a set of conjunctive queries involving from 1 to 30 relations. The queries have one selection on an attribute of each table. Each information source contains two relations and can perform remote operations. Therefore, the optimal plans involve pushing operations to be evaluated remotely at the sources.

The results of this experiment are shown in Figure 31. Figure 31(a) shows the planning time, in a logarithmic scale, for Sage, DP, Initial, PbR-FI, and PbR-SD as the query size grows. The times for PbR include both the generation of all the random initial plans and their rewriting. The times for Initial are the average of the initial plan construction across all restarts of each query. Sage is able to solve queries involving up to 6 relations, but larger queries cannot be solved within its search limit of 200,000 partial-plan nodes. DP scales better than Sage, but cannot solve queries of more than 9 relations in the 1000 second time limit. Both configurations of PbR scale better than Sage and DP. The first-improvement search strategy of PbR-FI is faster than the steepest descent of PbR-SD.

Figure 31(b) shows the cost of the query plans for the five planners. The cost for Initial is the average of the initial plans across all the restarts of each query. The plan cost is an estimate of the query execution cost. A logarithmic scale is used because of the increasingly larger absolute values of the plan costs for our conjunctive chain queries and the very high cost of the initial plans. PbR rewrites the very poor quality plans generated by Initial into high-quality plans. Both PbR and DP produce better plans than Sage (in the range tractable for Sage) for this experiment. This happens because they are searching the larger space of bushy query trees and can take greater advantage of parallel execution plans. PbR produces plans of quality comparable to DP for its tractable range and beyond that range PbR scales gracefully. The two configurations of PbR produce plans of similar cost, though PbR-FI needed less planning time than PbR-SD. PbR-SD generates all the plans in the local neighborhood in order to select the cheapest one, but PbR-FI only generates a portion of the neighborhood since it chooses the first plan of a cheaper cost, so PbR-FI is faster in average. Figure 31 shows empirically that in this domain the locally optimal moves of steepest descent do not translate in final solutions of a better cost than those produced by the first-improvement strategy.

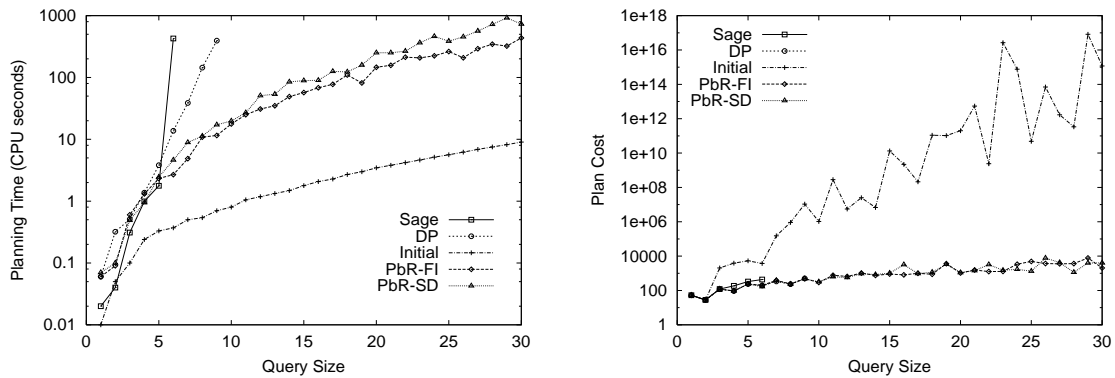

(a) Planning Time          (b) Plan Quality

Figure 31: Experimental Results: Distributed Query Planning

## 5. Related Work

In this section we review previous work related to the Planning by Rewriting framework. First, we discuss work on the disciplines upon which PbR builds, namely, classical AI planning, local search, and graph rewriting. Then, we discuss work related to our plan-rewriting algorithm.

## 5.1 AI Planning

PbR is designed to find a balance among the requirements of planning efficiency, high quality plans, flexibility, and extensibility. A great amount of work on AI Planning has focused on improving its average-case efficiency given that the general cases are computationally hard (Erol et al., 1995). One possibility is to incorporate domain knowledge in the form of search control. A recent example is TLPlan (Bacchus & Kabanza, 1995, 2000), a forward-search planner that has shown a remarkable scalability using control knowledge expressed in temporal logic. Some systems automatically learn search control for a given planning domain or even specific problem instances. Minton (1988b) shows how to deduce search control rules for a problem solver by applying explanation-based learning to problem-solving traces. He also discusses the impact of the utility problem. The utility problem, simply stated, says that the (computational) benefits of using the additional knowledge must outweigh the cost of applying it. PbR plan-rewriting rules also are subject to the utility problem. The quality improvement obtained by adding more rewriting rules to a PbR-based planner may not be worth the performance degradation. Another approach to automatically generating search control is by analyzing statically the operators (Etzioni, 1993) or inferring invariants in the planning domain (Gerevini & Schubert, 1998; Fox & Long, 1998; Rintanen, 2000). Abstraction provides yet another form of search control. Knoblock (1994a) presents a system that automatically learns abstraction hierarchies from a planning domain or a particular problem instance in order to speed up planning. plan-rewriting rules can be learned with techniques analogous to those used to learn search control. Ambite, Knoblock, & Minton (2000) present an approach to automatically learn the plan-rewriting rules based on comparing initial and optimal plans for example problems. Alternatively, analyzing the planning operators and which combinations of operators are equivalent with respect to the achievement of some goals can also lead to the automatic generation of the rewriting rules.

Local search algorithms have also been used to improve planning efficiency although in a somewhat indirect way. Planning can be reduced to solving a series of propositional satisfiability problems (Kautz & Selman, 1992). Thus, Kautz & Selman (1996) used an efficient satisfiability testing algorithm based on local search to solve the SAT encodings of a planning problem. Their approach proved more efficient than specialized planning algorithms. We believe that the power of their approach stems from the use of local search. PbR directly applies local search on the plan structures, as opposed to translating it first to a larger propositional representation.

Although all these approaches do improve the efficiency of planning, they do not specifically address plan quality, or else they consider only very simple cost metrics (such as the number of steps). Some systems learn search control that addresses both planning efficiency and plan quality (Estlin & Mooney, 1997; Borrajo & Veloso, 1997; Pérez, 1996). However, from the reported experimental results, PbR appears to be more scalable. Moreover, PbR provides an anytime algorithm while other approaches must run to completion.

## 5.2 Local Search

Local search has a long tradition in combinatorial optimization (Aarts & Lenstra, 1997; Papadimitriou & Steiglitz, 1982). Local improvement ideas have found application in many domains. Some of the general work most relevant to PbR is on constraint satisfaction, scheduling, satisfiability testing, and heuristic search.

In constraint satisfaction, local search techniques have been able to solve problems orders of magnitude more complex than the respective complete (backtracking) approaches. Minton et al. (Minton, Johnston, Philips, & Laird, 1990; Minton, 1992) developed a simple repair heuristic, min-conflicts, that could solve large constraint satisfaction and scheduling problems, such as the scheduling of operations in the Hubble Space Telescope. The min-conflicts heuristic just selects the variable value assignment that minimizes the number of constraints violated. This heuristic was used as the cost function of a gradient-descent search and also in an informed backtracking search.

In satisfiability testing a similar method, GSAT, was introduced by Selman, Levesque, & Mitchell (1992). GSAT solves hard satisfiability problems using local search where the repairs consist in changing the truth value of a randomly chosen variable. The cost function is the number of clauses satisfied by the current truth assignment. Their approach scales much better than the corresponding complete method (the Davis-Putnam procedure).

In work on scheduling and rescheduling, Zweben, Daun, & Deale (1994) define a set of general, but fixed, repair methods, and use simulated annealing to search the space of schedules. Our plans are networks of actions as opposed to their metric-time totally-ordered tasks. Also we can easily specify different rewriting rules (general or specific) to suit each domain, as opposed to their fixed strategies.

Our work is inspired by these approaches but there are several differences. First, PbR operates on complex graph structures (partial-order plans) as opposed to variable assignments. Second, our repairs are declaratively specified and may be changed for each problem domain, as opposed to their general but fixed repair strategies. Third, PbR accepts arbitrary measures of quality, not just constraint violations as in min-conflicts, or number of unsatisfied clauses as GSAT. Finally, PbR searches the space of valid solution plans, as opposed to the space of variable assignments which may be internally inconsistent.

Iterative repair ideas have also been used in heuristic search. Ratner & Pohl (1986) present a two-phase approach similar to PbR. In the first phase, they find an initial valid sequence of operators using an approximation algorithm. In the second phase, they perform local search starting from that initial sequence. The cost function is the plan length. The local neighborhood is generated by identifying segments in the current solution sequence and attempting to optimize them. The repair consists of a heuristic search with the initial state being the beginning of the segment and the goal the end of the segment. If a shorter path is found, the original sequence is replaced by the new shorter segment. A significant difference with PbR is that they are doing a state-space search, while PbR is doing a plan-space search. The least-committed partial-order nature of PbR allows it to optimize the plans in ways that cannot be achieved by optimizing linear subsequences.

## 5.3 Graph Rewriting

PbR builds on ideas from graph rewriting (Schürr, 1997). The plan-rewriting rules in PbR are an extension of traditional graph rewriting rules. By taking advantage of the semantics of planning PbR introduces partially-specified plan-rewriting rules, where the rules do not need to specify the completely detailed embedding of the consequent as in pure graph rewriting. Nevertheless, there are several techniques that can transfer from graph rewriting into Planning by Rewriting, particularly for fully-specified rules. Dorr (1995) defines an abstract machine for graph isomorphism and studies a set of conditions under which traditional graph rewriting can be performed efficiently. Perhaps a similar abstract machine for plan rewriting can be defined. The idea of rule programs also appears in this field and has been implemented in the PROGRES system (Schürr, 1990, 1997).

## 5.4 Plan Rewriting

The work most closely related to our plan-rewriting algorithm is plan merging (Foulser, Li, & Yang, 1992). Foulser et al. provide a formal analysis and algorithms for exploiting positive interactions within a plan or across a set of plans. However, their work only considers the case in which a set of operators can be replaced by *one* operator that provides the same effects to the rest of the plan and consumes the same or fewer preconditions. Their focus is on optimal and approximate algorithms for this type of operator merging. Plan rewriting in PbR can be seen as a generalization of operator merging where a subplan can replace another subplan. A difference is that PbR is not concerned with finding the optimal merge (rewritten plan) in a single pass of an optimization algorithm as their approach does. In PbR we are interested in generating possible plan rewritings during each rewriting phase, not the optimal one. The optimization occurs as the (local) search progresses.

Case-based planning (e.g., Kambhampati, 1992; Veloso, 1994; Nebel & Koehler, 1995; Hanks & Weld, 1995; Muñoz-Avila, 1998) solves a problem by modifying a previous solution. There are two phases in case-based planning. The first one identifies a plan from the library that is most similar to the current problem. In the second phase this previous plan is adapted to solve the new problem. PbR modifies a solution to the current problem, so there is no need for a retrieval phase nor the associated similarity metrics. Plan rewriting in PbR can be seen as a type of adaptation from a solution to a problem to an alternate solution for the same problem. That is, a plan rewriting rule in PbR identifies a pair of subplans (the replaced and replacement subplans) that may be interchangeable.

Veloso (1994) describes a general approach to case-based planning based on derivational analogy. Her approach works in three steps. First, the retrieval phase selects a similar plan from the library. Second, the parts of this plan irrelevant to the current problem are removed. Finally, her system searches for a completion of this plan selecting as much as possible the same decisions as the old plan did. In this sense the planning knowledge encoded in the previous solution is transferred to the generation of the new solution plan. The plan-rewriting algorithm for partially-specified rules of PbR can be seen as a strongly constrained version of this approach. In PbR the subplan in the rule consequent fixes the steps that can be added to repair the plan. We could use her technique of respecting previous choice points when completing the plan as a way of ensuring that most of the structure of the plan before and after the repair is maintained. This could be useful to constrain the number of rewritten plans for large rewriting rules.

Nebel and Koehler (1995) present a computational analysis of case-based planning. In this context they show that the worst-case complexity of plan modification is no better than plan generation and point to the limitations of reuse methods. The related problem in the PbR framework is the embedding of the replacement subplan for partially specified rules.

As we explained in Section 3.1.4 there may be pathological cases in which the number of embeddings is exponential in the size of the plan or deciding if the embedding exists is NP-hard. However, often we are not interested in finding all rewritings, for example when following a first improvement search strategy. In our experience the average case behavior seems to be much better as was presented in Section 4.

Systematic algorithms for case-based planning (such as Hanks & Weld, 1995) invert the decisions done in refinement planning to find a path between the solution to a similar old problem and the new problem. The rewriting rules in PbR indicate how to transform a solution into another solution plan based on domain knowledge, as opposed to the generic inversion of the refinement operations. Plan rewriting in PbR is done in a very constrained way instead of an open search up and down the space of partial plans. However, the rules in PBR may search the space of rewritings non systematically. Such an effect is ameliorated by using local search.

## 6. Discussion and Future Work

This paper has presented Planning by Rewriting, a new paradigm for efficient high-quality domain-independent planning. PbR adapts graph rewriting and local search techniques to the semantics of domain-independent partial-order planning. The basic idea of PbR consists in transforming an easy-to-generate, but possibly suboptimal, initial plan into a high-quality plan by applying declarative plan-rewriting rules in an iterative repair style.

There are several important advantages to the PbR planning approach. First, PbR is a *declarative domain-independent* framework, which brings the benefits of reusability and extensibility. Second, it addresses *sophisticated plan quality* measures, while most work in domain-independent planning has not addressed quality or does it in very simple ways. Third, PbR is *scalable* because it uses efficient local search methods. Finally, PbR is an *anytime* planning algorithm that allows balancing planning effort and plan quality in order to maximize the utility of the planning process.

Planning by Rewriting provides a domain-independent framework for local search. PbR accepts declarative domain specifications in an expressive operator language, declarative plan-rewriting rules to generate the neighborhood of a plan, complex quality metrics, inter-changeable initial plan generators, and arbitrary (local) search methods.

Planning by Rewriting is well suited to mixed-initiative planning. In mixed-initiative planning, the user and the planner interact in defining the plan. For example, the user can specify which are the available or preferred actions at the moment, change the quality criteria of interest, etc. In fact, some domains can only be approached through mixed-initiative planning. For example, when the quality metric is very expensive to evaluate, such as in geometric analysis in manufacturing, the user must guide the planner towards good quality plans in a way that a small number of plans are generated and evaluated. Another example is when the plan quality metric is multi-objective or changes over time. Several character-istics of PbR support mixed-initiative planning. First, because PbR offers complete plans, the user can easily understand the plan and perform complex quality assessment. Second, the rewriting rule language is a convenient mechanism by which the user can propose mod-ifications to the plans. Third, by selecting which rules to apply or their order of application the user can guide the planner.

Our framework achieves a balance between domain knowledge, expressed as plan-rewriting rules, and general local-search techniques that have proved useful in many hard combinatorial problems. We expect that these ideas will push the frontier of solvable problems for many practical domains in which high quality plans and anytime behavior are needed.

The planning style introduced by PbR opens several areas for future research. There is great potential for applying machine learning techniques to PbR. An important issue is the generation of the plan-rewriting rules. Conceptually, plan-rewriting rules arise from the chosen plan equivalence relation. All valid plans that achieve the given goals in a finite number of steps, i.e. all solution plans, are (satisfiability) equivalent. Each rule arises from a theorem that states that two subplans are equivalent for the purposes of achieving some goals, with the addition of some conditions that indicate in which context that rule can be usefully applied. The plan-rewriting rules can be generated by automated procedures. The methods can range from static analysis of the domain operators to analysis of sample equivalent plans that achieve the same goals but at different costs. Note the similarity with methods to automatically infer search control and domain invariants (Minton, 1988b; Etzioni, 1993; Gerevini & Schubert, 1998; Fox & Long, 1998; Rintanen, 2000), and also the need to deal with the utility problem. Ambite, Knoblock, & Minton (2000) present some results on learning plan rewriting rules based on comparing initial and optimal plans for sample problems.

Beyond learning the rewriting rules, we intend to develop a system that can automatically learn the optimal planner configuration for a given planning domain and problem distribution in a manner analogous to Minton's Multi-TAC system (Minton, 1996). Our system would perform a search in the configuration space of the PbR planner proposing candidate sets of rewriting rules and different search methods. By testing each proposed configuration against a training set of simple problems, the system would hill-climb in the configuration space in order to arrive at the most useful rewriting rules and search strategies for the given planning domain and distribution of problems.

There are many advanced techniques in the local search literature that can be adapted and extended in our framework. In particular, the idea of variable-depth rewriting leads naturally to the creation of rule programs, which specify how a set of rules are applied to a plan. We have already seen how in query planning we could find transformations that are better specified as a program of simple rewriting rules. For example, a sequence of `Join-Swap` transformations may put two retrieve operators on the same database together in the query tree and then `Remote-Join-Eval` would collapse the explicit join operator and the two retrieves into a single retrieval of a remote join. Cherniack & Zdonik (1996, 1998) present more complex examples of this sort of programs of rewriting rules in the context of a query optimizer for object-oriented databases.

As we discussed in Sections 3.1.3 and 3.1.4 the language of the antecedent of the rewriting rules can be more expressive than conjunctive queries while still remaining computationally efficient. For example, Figure 32 shows a rule from the manufacturing domain of Section 4.1 with a relationally-complete antecedent. This rule matches a subplan that contains a `spray-paint` operator, but does not contain either `punch` or `drill-press` operators that create holes of diameter smaller than 1 millimeter. In such case, the rule replaces the `spray-paint` operator by an `immersion-paint` operator. This rule would be useful in a situation in which painting by immersion could clog small holes.

```
(define-rule :name SP-by-IP-no-small-holes
 :if (:and (:operator ?n1 (spray-paint ?x ?c ?s))
           (:not (:and (:or (:operator ?n2 (punch ?x ?w ?o))
                            (:operator ?n3 (drill-press ?x ?w ?o)))
                       (:less ?w 1mm))))
 :replace (:operators (?n1))
 :with (:operator ?n4 (immersion-paint ?x ?c)))
```

Figure 32: Rule with a Relationally-Complete Antecedent

Another area for further research is the interplay of plan rewriting and plan execution. Sometimes the best transformations for a plan may only be known after some portion of the plan has been executed. This information obtained at run-time can guide the planner to select the appropriate rewritings. For example, in query planning the plans may contain information gathering actions (Ashish, Knoblock, & Levy, 1997) and depend on run-time conditions. This yields a form of dynamic query optimization. Interleaved planning and execution is also necessary in order to deal effectively with unexpected situations in the environment such as database or network failures.

An open area of research is to relax our framework to accept incomplete plans during the rewriting process. This expands the search space considerably and some of the benefits of PbR, such as its anytime property, are lost. But for some domains the shortest path of rewritings from the initial plan to the optimal may pass through incomplete or inconsistent plans. This idea could be embodied as a planning style that combines the characteristics of generative planning and Planning by Rewriting. This is reminiscent of the plan critics approach (Sacerdoti, 1975; Sussman, 1975). The resulting plan-rewriting rules can be seen as declarative specifications for plan critics. The plan refinements of both partial order planning (Kambhampati et al., 1995) and Hierarchical Task Network Planning (Erol, Nau, & Hendler, 1994) can be easily specified as plan-rewriting rules.

Applying PbR to other domains will surely provide new challenges and the possibility of discovering and transferring general planning techniques from one domain to another. We hope that the local-search methods used by PbR will help planning techniques to scale to large practical problems and conversely that the domain-independent nature of PbR will help in the analysis and principled extension of local search techniques.

## Acknowledgments

This paper is an extended version of (Ambite & Knoblock, 1997).

The research reported here was supported in part by a Fulbright/Ministerio of Educación y Ciencia of Spain scholarship, in part by the Defense Advanced Research Projects Agency (DARPA) and Air Force Research Laboratory, Air Force Materiel Command, USAF, under agreement number F30602-00-1-0504, in part by the National Science Foundation under grant number IRI-9313993, in part by the Rome Laboratory of the Air Force Systems Command and the Defense Advanced Research Projects Agency (DARPA) under contract numbers F30602-94-C-0210, F30602-97-2-0352, F30602-97-2-0238, F30602-98-2-0109, in part by the United States Air Force under contract number F49620-98-1-0046, and in part by the Integrated Media Systems Center, a National Science Foundation Engineering Research

Center, Cooperative Agreement No. EEC-9529152. The U.S.Government is authorized to reproduce and distribute reports for Governmental purposes notwithstanding any copyright annotation thereon. The views and conclusions contained herein are those of the authors and should not be interpreted as necessarily representing the official policies or endorsements, either expressed or implied, of any of the above organizations or any person connected with them.